\documentclass[acmsmall,screen]{acmart}
\AtBeginDocument{%
  \providecommand\BibTeX{{%
    \normalfont B\kern-0.5em{\scshape i\kern-0.25em b}\kern-0.8em\TeX}}}

\setcopyright{acmcopyright}
\copyrightyear{2018}
\acmYear{2018}
\acmDOI{10.1145/1122445.1122456}

\acmJournal{JACM}
\acmVolume{37}
\acmNumber{4}
\acmArticle{111}
\acmMonth{8}

\usepackage{amsmath}
\usepackage{subfigure}
\usepackage{url}            
\usepackage{booktabs}       
\usepackage{amsfonts}       
\usepackage{nicefrac}       
\usepackage{microtype}      
\usepackage{xcolor}
\usepackage{multirow}
\usepackage{wrapfig}
\usepackage{array}
\usepackage{soul}
\usepackage{float}
\usepackage{caption}
\usepackage{bm}
\usepackage{color}

\usepackage{amsthm}  
\usepackage{wasysym}
\usepackage[ruled,vlined,algo2e]{algorithm2e}

\usepackage{color}
\usepackage{dsfont}	
\usepackage{footnote}
\usepackage{epstopdf}
\usepackage{enumitem}
\usepackage{comment}
\usepackage{multirow}
\usepackage{wrapfig}

\usepackage{booktabs}       
\usepackage{amsfonts}       
\usepackage{nicefrac}       
\usepackage{microtype}      
\usepackage{amsmath}
\usepackage{adjustbox}
\usepackage[flushleft]{threeparttable}

\begin{document}

\title{Deep Learning-Based  Human Pose Estimation: A Survey}

\author{Ce Zheng}
\authornote{Both authors contributed equally to this research.}
\email{cezheng@knights.ucf.edu}
\affiliation{%
  \institution{University of Central Florida}
  \streetaddress{4328 Scorpius St, Suite 245}
  \city{Orlando}
  \state{Florida}
  \country{USA}
  \postcode{32816}
}

\author{Wenhan Wu}
\authornotemark[1]
\email{wwu25@uncc.edu}
\affiliation{%
  \institution{University of North Carolina at Charlotte}
  \streetaddress{9201 University City Blvd}
  \city{Charlotte}
  \state{North Carolina}
  \country{USA}
  \postcode{28223}}

\author{Chen Chen}
\email{chen.chen@crcv.ucf.edu}
\affiliation{%
  \institution{University of Central Florida}
  \streetaddress{4328 Scorpius St, Suite 245}
  \city{Orlando}
  \state{Florida}
  \country{USA}
  \postcode{32816}
}

\author{Taojiannan Yang}
\email{taoyang1122@knights.ucf.edu}
\affiliation{%
  \institution{University of Central Florida}
  \streetaddress{4328 Scorpius St, Suite 245}
  \city{Orlando}
  \state{Florida}
  \country{USA}
  \postcode{32816}
}

\author{Sijie Zhu}
\email{sizhu@knights.ucf.edu}
\affiliation{%
  \institution{University of Central Florida}
  \streetaddress{4328 Scorpius St, Suite 245}
  \city{Orlando}
  \state{Florida}
  \country{USA}
  \postcode{32816}
}

\author{Ju Shen}
\email{jshen1@udayton.edu}
\affiliation{%
  \institution{University of Dayton}
  \streetaddress{300 College Park}
  \city{Dayton}
  \state{Ohio}
  \country{USA}
  \postcode{45469}}

\author{Nasser Kehtarnavaz}
\email{kehtar@utdallas.edu}
\affiliation{%
  \institution{University of Texas at Dallas}
  \streetaddress{ 800 W. Campbell Road}
  \city{Richardson}
  \state{Texas}
  \country{USA}
  \postcode{75080}}

\author{Mubarak Shah}
\email{shah@crcv.ucf.edu}
\affiliation{%
  \institution{University of Central Florida}
  \streetaddress{4328 Scorpius St, Suite 245}
  \city{Orlando}
  \state{Florida}
  \country{USA}
  \postcode{32816}
}

\renewcommand{\shortauthors}{Zheng and Wu, et al.}

\begin{abstract}
 Human pose estimation aims to locate the human body parts and build human body representation (e.g., body skeleton) from input data such as images and videos. It has drawn increasing attention during the past decade and has been utilized in a wide range of applications including human-computer interaction, motion analysis, augmented reality, and virtual reality. Although the recently developed deep learning-based solutions have achieved high performance in human pose estimation, there still remain challenges due to insufficient training data, depth ambiguities, and occlusion. 
The goal of this survey paper is to provide a comprehensive review of recent deep learning-based solutions  for both 2D and 3D pose estimation  via a systematic analysis and comparison of these solutions based on their input data and inference procedures. More than {260} research papers since 2014 are covered in this survey. Furthermore, 2D and 3D human pose estimation datasets and evaluation metrics are included. Quantitative performance comparisons of the reviewed methods on popular datasets are summarized and discussed. Finally, the challenges involved, applications, and future research directions are concluded. A regularly updated project page is provided:
\url{https://github.com/zczcwh/DL-HPE}
\end{abstract}

\ccsdesc[500]{Computing methodologies~Computer vision}
\ccsdesc[500]{General and reference~Surveys and overviews}

\keywords{Survey of human pose estimation, 2D and 3D pose estimation, deep learning-based pose estimation, pose estimation datasets, pose estimation metrics}

\maketitle

\section{Introduction}\label{sec:introduction}

Human pose estimation (HPE), which has been extensively studied in computer vision literature,  involves estimating  the configuration of human body parts from input data captured by sensors, in particular images and videos.  HPE provides geometric and motion information about the human body which has been applied to a wide range of applications (e.g., human-computer interaction, motion analysis, augmented reality (AR), virtual reality (VR), healthcare, etc.). With the rapid development of deep learning solutions in recent years, such solutions have been shown to outperform classical computer vision methods in various tasks including image classification \cite{krizhevsky2012imagenet}, semantic segmentation \cite{Semantic_Segmentation}, and object detection \cite{ren2015faster}. 
Significant progress and remarkable performance have already been made by  employing  deep  learning  techniques  in  HPE tasks. However, challenges such as occlusion, insufficient training data, and depth ambiguity still pose difficulties to be overcome.  2D HPE from  images and videos with 2D pose annotations is easily achievable and high performance has been reached for the  human  pose  estimation of  a single person using deep learning techniques. More recently, attention has been paid to highly occluded multi-person HPE in complex scenes. In contrast, for 3D HPE, obtaining  accurate  3D  pose  annotations  is  much  more difficult than its 2D counterpart. Motion capture systems can collect 3D pose annotation in controlled lab environments;  however, they  have limitations for in-the-wild environments. For 3D HPE from monocular RGB images and videos, the main challenge is depth ambiguities. In a multi-view setting, viewpoints association is the key issue that needs to be addressed. Some works have utilized sensors such as depth sensors, inertial measurement units (IMUs), and radio frequency devices, but these approaches are usually not cost-effective and require special-purpose hardware. 

Given the rapid progress in HPE research, this article attempts to track recent advances and summarize their achievements in order to provide 
a clear picture of current research on deep learning-based 2D and 3D HPE.

\begin{figure}[htp]
\vspace{-0.20cm}
  \centering
  \includegraphics[width=1\linewidth]{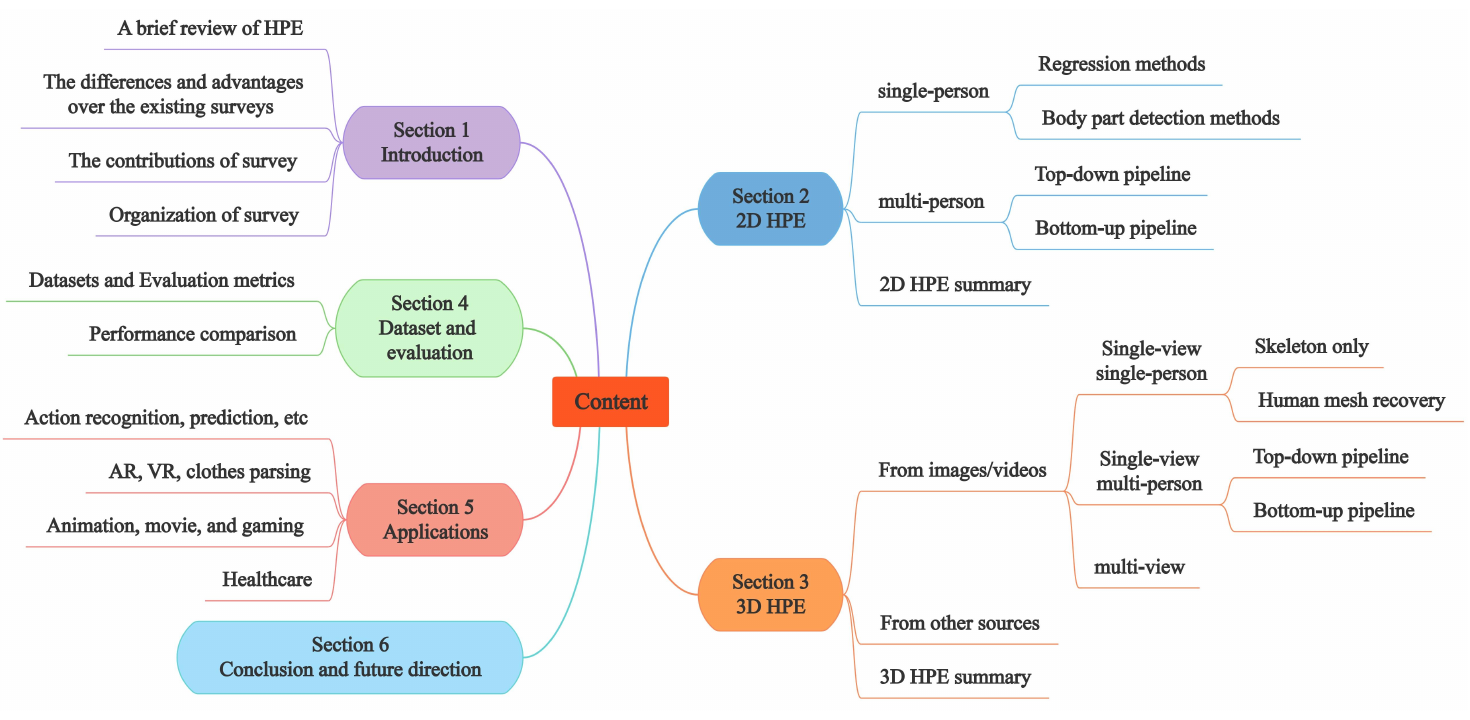}
  \vspace{-10pt}
  \caption{Taxonomy of this survey.}
  \vspace{-5pt}
    \label{fig:Taxonomy}
  \vspace{-10pt}
\end{figure}


\subsection{Previous surveys and our contributions}
There are several related surveys and reviews previously reported on HPE. Among them, \cite{moeslund2001survey,moeslund2006survey,poppe2007vision,ji2009advances} focus on the general field of visual-based human motion capture including pose estimation, tracking, and action recognition. Therefore, pose estimation is only one of the topics covered in these surveys. The research works on 3D HPE before 2012 are reviewed in \cite{holte2012human}. The body parts parsing-based methods for single-view and multi-view HPE are reported in \cite{liu2015survey}. These surveys published during 2001-2015 mainly focused  on  conventional methods without deep learning.
A survey on both traditional and deep learning-based methods related to HPE is presented in \cite{gong2016human}. However, only a handful of deep learning-based approaches are included. The survey in \cite{sarafianos20163d} covers 3D HPE methods with RGB inputs, while the survey in \cite{munea2020progress} only reviews 2D HPE methods. Monocular HPE from the classical to recent deep learning-based methods (till 2019, less than 100 papers) is summarized in \cite{chen2020monocular}. However, it only covers 2D HPE and 3D single-view HPE from monocular cameras. 3D multi-view HPE from monocular cameras and 3D HPE from other sensors are ignored. Also, no extensive performance comparisons or in-depth analyses are given, and the discussion on existing challenges and future directions is relatively short. 

\textit{This survey aims to address the shortcomings of the previous surveys in terms of providing a systematic review of the recent deep learning-based solutions to 2D and 3D HPE but also covering other aspects of HPE including the performance evaluation of (2D and 3D) HPE methods on popular datasets, their applications, and comprehensive discussion. The key points that distinguish this survey from the previous ones are as follows:}
\setlist{nolistsep}
\begin{itemize}[noitemsep,leftmargin=*]
\item A comprehensive review of recent deep learning-based 2D and 3D HPE methods (up to 2022 with more than 260 papers) is provided by categorizing them according to 2D or 3D scenarios, single-view or multi-view, from monocular images/videos or other sources, and learning paradigm. 

\item Extensive performance evaluation of 2D and 3D HPE methods. We summarize and compare reported performances of promising methods on common datasets based on their categories. The comparison of results provides cues for the strengths and weaknesses of different methods, revealing the research trends and future directions of HPE.

\item An overview of a wide range of HPE applications, such as surveillance, AR/VR, and healthcare.   

\item An thorough discussion of 2D and 3D HPE is presented in terms of key challenges in HPE pointing to potential future research toward improving  performance.

\end{itemize}


\subsection{Paper organization}


HPE is divided into two main categories: 2D HPE (\textsection~\ref{2D}) and 3D HPE (\textsection~\ref{3D}). Fig. \ref{fig:Taxonomy} shows the taxonomy of deep learning methods for HPE.
According to the number of people, 2D HPE methods are categorized into single-person and multi-person settings. 
For single-person methods (\textsection~\ref{2DsingleHPE}), there are two categories: regression methods and
heatmap-based methods.
For multi-person methods (\textsection~\ref{2D multi-person}), there are also two types of methods: top-down methods and
bottom-up methods.

3D HPE methods are classified according to the input source types: monocular RGB images and videos (\textsection~\ref{3drgb}), or other sensors (e.g., inertial  measurement  unit  sensors, \textsection~\ref{3dother}).  The majority of these methods use monocular RGB images and videos, and they are further   divided  into  single-view single-person (\textsection~\ref{3d1view1person}); single-view multi-person (\textsection~\ref{3d1viewnperson});  and  multi-view methods (\textsection~\ref{3dmultiview}). Multi-view settings are deployed mainly for multi-person pose estimation. Hence, single-person or multi-person is not specified in this category. 

Next, depending on the 2D and 3D HPE pipelines, the datasets and evaluation metrics commonly used are summarized followed by a comparison of results of the promising methods (\textsection~\ref{dataset}). In addition, various  applications of HPE such as AR/VR are mentioned (\textsection~\ref{Application}). Finally, the paper ends by an 
thorough discussion of 
some promising directions for future research
(\textsection~\ref{Discussion}).

\section{2D human pose estimation}\label{2D}

2D HPE methods estimate the 2D position or spatial location of human body keypoints from images or videos. Traditional 2D HPE methods adopt different  hand-crafted feature extraction techniques 
for body parts, and these early works describe the human body as a stick figure to obtain global pose structures. Recently, deep learning-based approaches have achieved a major breakthrough in HPE by improving the results significantly.  In the following, we review deep learning-based 2D HPE methods with respect to single-person and multi-person scenarios. 

\subsection{2D single-person pose estimation}  \label{2DsingleHPE}
\begin{figure}[htp]
  \centering
  \includegraphics[width=0.8\linewidth]{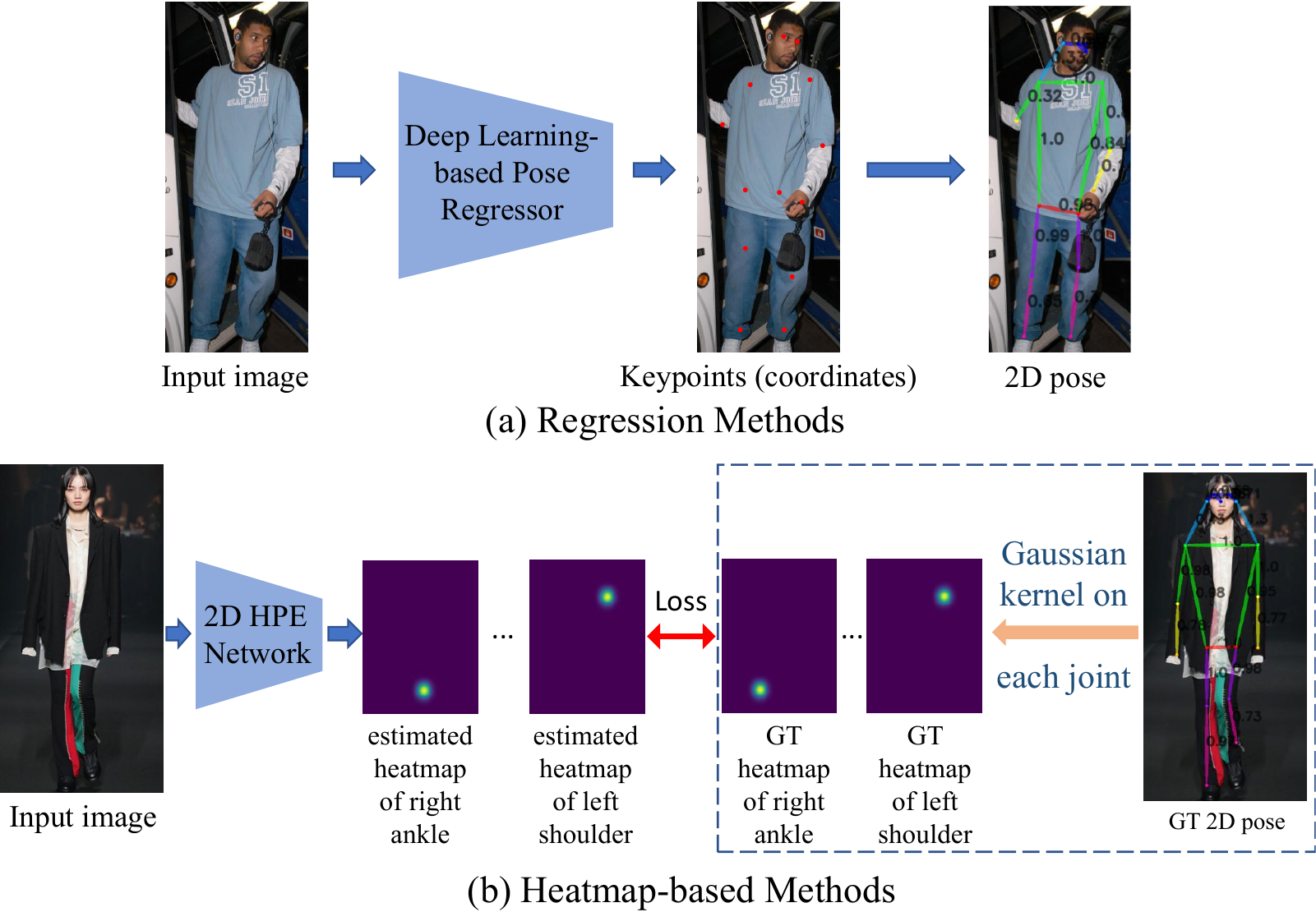}
\vspace{-0.25cm}
   \caption{\footnotesize{Single-person 2D HPE frameworks. (a) Regression methods directly learn a mapping (via a deep neural network) from the original image to the  kinematic  body  model  and  produce  joint  coordinates. (b) Given the ground-truth 2D pose, the ground-truth heatmaps of each joint are generated by applying a Gaussian kernel to each joint's location. Then, heatmap-based methods utilize a model to predict the heatmap of each joint. }} 
  \label{fig:2dsingleframeworks}
  \vspace{-10pt}
\end{figure}

2D single-person pose estimation is used to localize human body joint positions when the input is a single-person image. If there are several people, the input image is cropped first so that there is only one person in each cropped patch (or sub-image). This process can be achieved automatically by an upper-body detector \cite{micilotta2006real} or a full-body detector \cite{ren2015faster}.
In general, there are two categories for single-person pipelines that employ deep learning techniques: regression methods and heatmap-based methods. Regression methods apply an end-to-end framework to learn a mapping from the input image to the positions of body joints or parameters of human body models \cite{toshev2014deeppose}. The goal of heatmap-based methods is to predict approximate locations of body parts and joints \cite{chen2014articulated} \cite{newell2016stacked}, which are supervised by heatmaps representation \cite{tompson2015efficient,wei2016convolutional}. Heatmap-based frameworks are now widely used in 2D HPE tasks. The general frameworks of 2D single-person HPE methods are depicted in Fig. \ref{fig:2dsingleframeworks}.


\subsubsection{Regression methods}
There are many works based on the \textbf{regression framework} (e.g., \cite{toshev2014deeppose,pfister2014deep,carreira2016human,sun2017compositional,luvizon2019human,nibali2018numerical,li2014heterogeneous,fan2015combining,luvizon20182d,zhang2019fast,li2021pose,panteleris2021pe,mao2021tfpose,mao2022poseur}) to predict joint coordinates from images as shown in Fig.  \ref{fig:2dsingleframeworks} (a). Using AlexNet \cite{krizhevsky2012imagenet} as the backbone, Toshev and Szegedy \cite{toshev2014deeppose} proposed a cascaded deep neural network regressor named DeepPose to learn keypoints from images. Due to the impressive performance of DeepPose, the research paradigm of HPE began to shift from classic approaches to deep learning, in particular convolutional neural networks (CNNs). 
Sun et al. \cite{sun2017compositional} introduced a structure-aware regression method called "compositional pose regression" based on ResNet-50 \cite{resnet}. This method adopts a re-parameterized and bone-based representation that contains human body information and pose structure, instead of the traditional joint-based representation. Luvizon et al. \cite{luvizon2019human} proposed an end-to-end regression approach for HPE using soft-argmax function to convert feature maps into joint coordinates in a fully differentiable framework. Li et al. \cite{li2021pose} first designed a transformer-based cascade network for regressing human keypoints. The spatial correlation of joints and appearance is captured by self-attention mechanism. Different from the previous methods, Li et al. \cite{li2021human} proposed a normalizing flow model named RLE (Log-likelihood Estimation) to capture the distribution of joint location, aiming for finding the optimized parameters by residual log-likelihood estimation.




A good feature that encodes rich pose information is critical for regression-based methods. One popular strategy to learn better feature representation is \textbf{multi-task learning} \cite{multitask}. By sharing representations between related tasks (e.g., pose estimation and pose-based action recognition), the model can generalize better on the original task (pose estimation).
Following this direction, Li et al. \cite{li2014heterogeneous} proposed a heterogeneous multi-task framework that consists of two tasks: predicting joint coordinates from full images by a regressor and detecting body parts from image patches using a sliding window. Fan et al. \cite{fan2015combining} proposed a dual-source (i.e., image patches and full images) CNN for two tasks: joint detection which determines whether a patch contains a body joint, and joint localization which finds the exact location
of the joint in the patch. Each task corresponds to a loss function, and the combination of two tasks leads to improved results.


\subsubsection{Heatmap-based methods}
Instead of estimating the 2D coordinates of human joints directly, heatmap-based methods for HPE aim to estimate the 2D heatmaps which are generated by adding 2D Gaussian kernels on each joint's location as shown in Fig. \ref{fig:2dsingleframeworks}. 
Concretely, the goal is to estimate $K$ heatmaps $\{H_1, H_2,..., H_K\}$ for a total of $K$ keypoints. The pixel value $H_i(x,y)$ in each keypoint heatmap indicates the probability that the keypoint lies in the position $(x,y)$ (see Fig. \ref{fig:2dsingleframeworks} (b)). The target (or ground-truth) heatmap is generated by a 2D Gaussian centered at the ground-truth joint location \cite{tompson2015efficient}\cite{tompson2014joint}. Thus pose estimation networks are trained by minimizing the discrepancy (e.g., the Mean Squared-Error (MSE)) between the predicted heatmaps and target heatmaps. Compared with joint coordinates, heatmaps preserve the spatial location information while it can make the training process smoother.  

Therefore, there is a recent growing interest in leveraging heatmaps to represent the joint locations and developing \textbf{effective CNN architectures} for HPE, e.g.,  \cite{tompson2014joint,ramakrishna2014pose,tompson2015efficient,lifshitz2016human,bulat2016human,newell2016stacked,wei2016convolutional,gkioxari2016chained,belagiannis2017recurrent,yang2017learning,luo2018lstm,debnath2018adapting,Xiao_2018_ECCV,zhang2019human,artacho2020unipose, li2022simcc}.
As one of the fundamental works, Wei et al. \cite{wei2016convolutional} introduced a convolutional networks-based sequential framework named Convolutional Pose Machines (CPM) to predict the locations of keypoints with multi-stage processing (the convolutional networks in each stage utilize the 2D belief maps generated from previous stages and produce the increasingly refined predictions of body part locations). At the same time, Newell et al. \cite{newell2016stacked} proposed an encoder-decoder network named "stacked hourglass" to repeat bottom-up and top-down processing with intermediate supervision. In this work, the encoder squeezes features through the bottleneck and then the decoder expands them for the substage.
The stacked hourglass (SHG) network consists of consecutive steps of pooling and upsampling layers to capture information at every scale. Since then, complex variations of the SHG architecture were developed for HPE.  Following \cite{newell2016stacked}, Chu et al. \cite{chu2017multi} designed novel Hourglass Residual Units (HRUs), which extend the residual units with a side branch of filters with larger receptive fields, to capture features from various scales. Yang et al. \cite{yang2017learning} designed a multi-branch Pyramid Residual Module (PRM) to replace the residual unit in SHG, leading to enhanced invariance in scales of deep CNNs. Sun et al. \cite{sun2019deep} presented a novel High-Resolution Net (HRNet) to learn reliable high-resolution representations by connecting multi-resolution subnetworks in parallel and conducting repeated multi-scale fusions, which results in more accurate keypoint heatmap prediction. Inspired by HRNet, Yu et al. \cite{yu2021lite} introduced a light-weighted HRNet named Lite-HRNet, which designed conditional channel weighting blocks to exchange information between channels and resolutions. Recently, due to the superior performance, the HRNet \cite{sun2019deep} and its variations \cite{cheng2020higherhrnet,yuan2021hrformer,yu2021lite} have been widely adopted in HPE and other pose-related tasks. 


With the emergence of \textbf{Generative Adversarial Networks (GANs)} \cite{gans}, they are explored in HPE to generate biologically plausible pose configurations and to discriminate the predictions with high confidence from those with low confidence, which could infer the poses of the occluded body parts. Inspired by hourglass architecture which efficiently refines joints, Chen et al. \cite{chen2017adversarial} constructed a structure-aware conditional adversarial network--Adversarial PoseNet--which contains an hourglass network-based pose generator and two discriminators to discriminate reasonable body poses from unreasonable ones. Chou et al. \cite{chou2018self} built an adversarial learning-based network with two stacked hourglass networks sharing the same structure as the discriminator and generator, respectively. The generator estimates the location of each joint, and the discriminator distinguishes between the ground-truth heatmaps and predicted ones. Unlike GANs-based methods that take the HPE network as the generator and utilize the discriminator to provide supervision, Peng et al. \cite{peng2018jointly} developed an adversarial data augmentation network to optimize data augmentation and network training by treating the HPE network as a discriminator and using augmentation network as a generator to perform adversarial augmentations.

Besides these efforts in designing effective networks for HPE, \textbf{body structure information} is also investigated to provide more and better supervision information for building HPE networks. Yang et al. \cite{yang2016end} designed an end-to-end CNN framework for HPE, which can find hard negatives by incorporating the spatial and appearance consistency among human body parts. A structured feature-level learning framework was proposed in \cite{chu2016structured} for reasoning the correlations among human body joints in HPE, which captures richer information of human body joints and improves the learning results. 
Ke et al. \cite{ke2018multi} designed a multi-scale structure-aware neural network, which combines multi-scale supervision, multi-scale feature combination, structure-aware loss information scheme, and a keypoint masking training method to improve HPE models in complex scenarios. Tang et al. \cite{tang2018deeply} built an hourglass-based supervision network, termed as Deeply Learned Compositional Model, to describe the complex and realistic relationships among body parts and learn the compositional pattern information (the orientation, scale, and shape information of each body part) in human bodies. Different from the previous approaches which consider all body parts, Tang and Wu \cite{tang2019does} revealed that not all parts are related to each other, therefore introducing a Part-based Branches Network to learn representations specific to each part group rather than a shared representation for all parts. 

\textbf{Human poses in video sequences} are (3D) spatio-temporal signals. Therefore, modeling spatio-temporal information is important for HPE from videos. Jain et al. \cite{jain2014modeep} designed a two-branch CNN framework to incorporate both color and motion features within frame pairs to build an expressive temporal-spatial model in HPE. 
Pfister et al. \cite{pfister2015flowing} proposed a CNN that can utilize temporal context information from multiple frames by using optical flow to align predicted heatmaps from neighboring frames. Different from the previous video-based methods which are computationally intensive, Luo et al. \cite{luo2018lstm} introduced a recurrent structure with Long Short-Term Memory to capture temporal geometric consistency and dependency from different frames. This method results in faster training time for the HPE network for videos. Zhang et al. \cite{zhang2020key} introduced a keyframe proposal network for capturing spatial and temporal information from frames, and a human pose interpolation module for efficient video-based HPE. 

\subsection{\text{2D multi-person pose estimation}}\label{2D multi-person}
\begin{figure}[htp]
  \centering
  \includegraphics[width=0.8\linewidth]{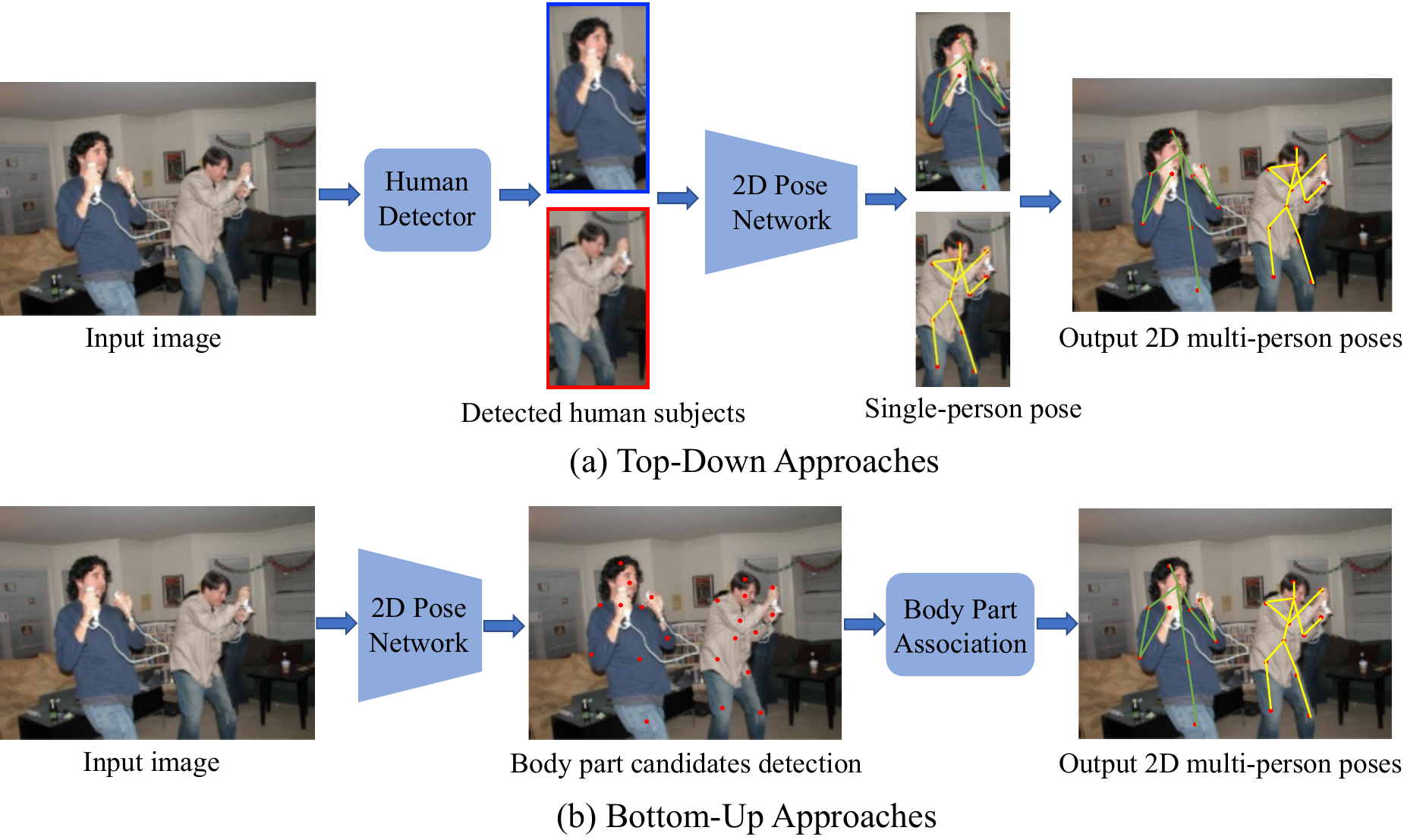}
\vspace{-0.25cm}
   \caption{\footnotesize{Illustration of the multi-person 2D HPE frameworks. (a) Top-down approaches have two sub-tasks: (1) human detection and (2) pose estimation in the region of a single human; (b) Bottom-up approaches also have two sub-tasks: (1) detect all keypoints candidates of body parts and (2) associate body parts in different human bodies and assemble them into individual pose representations.}}
   \vspace{-5pt}
  \label{fig:2dmultiframeworks}
  \vspace{-5pt}
\end{figure}

Compared to single-person HPE, multi-person HPE is more difficult and challenging because it needs to figure out the number of people and their positions, and how to group keypoints for different people. In order to solve these problems, multi-person HPE methods can be classified into top-down and bottom-up methods. Top-down methods employ off-the-shelf person detectors to obtain a set of boxes (each corresponding to one person) from the input images and then apply single-person pose estimators to each person box to generate multi-person poses. Different from top-down methods, bottom-up methods locate all the body joints in one image first and then group them into individual subjects. In the top-down pipeline, the number of people in the input image will directly affect the computing time. The computing speed for bottom-up methods is usually faster than top-down methods since they do not need to detect the pose for each person separately. Fig. \ref{fig:2dmultiframeworks} shows the general frameworks for 2D multi-person HPE methods.


\subsubsection{Top-down pipeline}
In the top-down pipeline as shown in Fig. \ref{fig:2dmultiframeworks} (a), there are two important parts: a human body detector to obtain person bounding boxes and a single-person pose estimator to predict the locations of keypoints within these bounding boxes. A line of works focus on \textbf{designing and improving the modules} in HPE networks, e.g., \cite{Papandreou_2017_CVPR,huang2017coarse,Xiao_2018_ECCV,sun2019deep,li2019rethinking,moon2019posefix,wang2020graph,huang2020devil,cai2020learning,zhang2020distribution, liu2021polarized}. To answer the question "how good could a simple method be" in building an HPE network, Xiao et al. \cite{Xiao_2018_ECCV} added a few deconvolutional layers in the ResNet (backbone network) to build a simple yet effective structure to produce heatmaps for high-resolution representations. 
To improve the accuracy of keypoint localization, Wang et al. \cite{wang2020graph} introduced a two-stage graph-based and model-agnostic framework, called Graph-PCNN. It consists of a localization subnet to obtain rough keypoint locations and a graph pose refinement module to get refined keypoints localization representations. 
Cai et al. \cite{cai2020learning} introduced a multi-stage network with a Residual Steps Network (RSN) module to learn delicate local representations by efficient intra-level feature fusion strategies, and a Pose Refine Machine (PRM) module to find a trade-off between local and global representations in the features. 


\textbf{Estimating poses under occlusion and truncation scenes} often occurs in multi-person settings since the overlapping of limbs is inevitable. 
Human detectors may fail in the first step of top-down pipeline due to occlusion. Thus, robustness to occlusion or truncation is an important aspect of the multi-person HPE approaches. Towards this goal, Iqbal and Gall \cite{iqbal2016multi} built a convolutional pose machine-based pose estimator to estimate the joint candidates. Then they used integer linear programming to solve the joint-to-person association problem and obtain human body poses even in presence of severe occlusions. Fang et al. \cite{fang2017rmpe} designed a regional multi-person pose estimation (RMPE) approach to improve the performance of HPE in complex scenes. The RMPE framework has three parts: Symmetric Spatial Transformer Network (to detect single person region within an inaccurate bounding box), Parametric Pose Non-Maximum-Suppression (to solve the redundant detection problem), and Pose-Guided Proposals
Generator (to augment training data). Papandreou et al. \cite{Papandreou_2017_CVPR} proposed a two-stage architecture with a Faster R-CNN person detector to create bounding boxes for candidate human bodies and a keypoint estimator to predict the locations of keypoints by using a form of heatmap-offset aggregation. The method works well in occluded and cluttered scenes. To alleviate the occlusion problem in HPE, Chen et al. \cite{chen2018cascaded} presented a Cascade Pyramid Network (CPN) which includes two parts: GlobalNet (a feature pyramid network to predict the invisible keypoints) and RefineNet (a network to integrate all levels of features from the GlobalNet with a keypoint mining loss). Their results reveal that CPN has a good performance in predicting occluded keypoints. 
Su et al. \cite{su2019multi} designed two modules, the Channel Shuffle Module and the Spatial \& Channel-wise Attention Residual Bottleneck, to achieve channel-wise and spatial information enhancement for better multi-person HPE under occluded scenes.
Qiu et al. \cite{qiu2020peeking} developed an Occluded Pose Estimation and Correction module and an occluded pose dataset to solve the occlusion problem in crowd pose estimation.
Umer et al. \cite{umer2020self} proposed a keypoint correspondence framework to recover missed poses using temporal information of the previous frame in occluded scenes. The network is trained using self-supervision to improve the pose estimation results on sparsely annotated video datasets.  

Recently, \textbf{transformer-based methods} have attracted more attention \cite{li2021test,Li_2021_ICCV,yang2021transpose,yuan2021hrformer, shi2022end, ma2022ppt} since the attention modules in transformer can obtain long-range dependencies and global evidence of the predicted keypoints, which are more powerful than CNNs. The early exploration \cite{yang2021transpose} proposed a transformer-based model for 2D HPE named TransPose, which utilizes the attention layers to predict the heatmaps of the keypoints and learn the fine-grained evidence for HPE in occlusion scenarios. Following \cite{yang2021transpose}, Li et al. \cite{Li_2021_ICCV} built a pure transformer-based model named TokenPose to capture the constraint cues and visual appearance relationships by using token representation. In contrast to the methods based on the  vision transformer which learn the representations in low resolution, Yuan et al. \cite{yuan2021hrformer} presented a high-resolution transformer named HRFormer by exchanging the blocks in HRNet \cite{cheng2020higherhrnet} with transformer modules, which improves the memory and computing efficiency.
Ma et al. \cite{ma2022ppt} applied the token-Pruned Pose Transformer (PPT) for locating the areas of the human body, which enables the model to estimate the multi-view pose efficiently. Different from the traditional two-step structures in HPE, Shi et al. \cite{shi2022end} proposed a fully end-to-end framework based on the attention mechanism, which directly estimates the instance-aware body poses. 

Besides the image-based works introduced above, \textbf{multi-frame pose estimation in videos} is also popular in multi-person 2D HPE \cite{guo2018multi, bertasius2019learning, liu2021deep, xu2021vipnas, liu2022temporal}, which leverages the temporal information in video sequences to facilitate the pose estimation. In order to reduce the cost of labeling frames in the video, Bertasius et al.\cite{bertasius2019learning} proposed a network named PoseWarper, which improves the label propagation between frames and benefits the training with the sparse annotations. To alleviate the motion blur and pose occlusions among video frames, Liu et al. \cite{liu2021deep} designed a network named DCpose for multi-frame HPE, which contains three modules (Pose Temporal Merger, Pose Residual Fusion, and Pose Correction Network) to exploit the temporal information between frames for keypoint detection. Nevertheless, these methods failed to fully utilize the information from neighboring frames. To solve this issue, Liu et al. \cite{liu2022temporal} introduced a hierarchical alignment framework for alleviating the aggregation of unaligned contexts between two frames.


\subsubsection{Bottom-up pipeline}
The bottom-up pipeline (e.g., \cite{insafutdinov2017arttrack,cao2017realtime,newell2017associative,fieraru2018learning,tian2019directpose,kreiss2019pifpaf,Nie_2019_ICCV,jin2020differentiable,cheng2020higherhrnet, wang2022lite, wang2022regularizing}) has two main steps including body joint detection (i.e., extracting local features and predicting body joint candidates) and joint candidates assembling for individual bodies (i.e., grouping joint candidates to build pose representations with part association strategies) as shown in Fig. \ref{fig:2dmultiframeworks} (b). 

Pishchulin et al. \cite{pishchulin2016deepcut} proposed a Fast R-CNN-based body part detector named DeepCut, which is one of the earliest \textbf{two-stage} bottom-up approaches. It first detects all the body part candidates, then labels each part and assembles these parts using integer linear programming (ILP) to a final pose. However, DeepCut is computationally expensive. To this end, Insafutdinov et al.\cite{insafutdinov2016deepercut} introduced DeeperCut to improve DeepCut by applying a stronger body part detector with a better incremental optimization strategy and image-conditioned pairwise terms to group body parts, leading to improved performance as well as a faster speed. Later, Cao et al. \cite{cao2017realtime} built a detector named OpenPose, which uses Convolutional Pose Machines \cite{wei2016convolutional} to predict keypoint coordinates via heatmaps and  Part Affinity Fields (PAFs, a set of 2D vector fields with vector maps that encode the position and orientation of limbs) to associate the keypoints to each person. OpenPose largely accelerates the speed of bottom-up multi-person HPE. Based on the OpenPose framework, Zhu et al. \cite{zhu2017multi} improved the OpenPose structure by adding redundant edges to increase the connections between joints in PAFs and obtained better performance than the baseline approach. Although OpenPose-based methods have achieved impressive results on high-resolution images, they have poor performance with low-resolution images and occlusions. To address this problem, Kreiss et al. \cite{kreiss2019pifpaf} proposed a bottom-up method called PifPaf that uses a Part Intensity Field to predict the locations of body parts and a Part Association Field to represent the joints association. This method outperformed previous OpenPose-based approaches on low-resolution and occluded scenes. 
Motivated by OpenPose \cite{cao2017realtime} and stacked hourglass structure  \cite{newell2016stacked}, Newell et al. \cite{newell2017associative} introduced a \textbf{single-stage} deep network to simultaneously obtain pose detection and group assignments. Following \cite{newell2017associative}, Jin et al. \cite{jin2020differentiable} proposed a new differentiable Hierarchical Graph Grouping method to learn the human part grouping. Based on \cite{newell2017associative} and \cite{sun2019deep}, Cheng et al. \cite{cheng2020higherhrnet} proposed an extension of HRNet, named Higher Resolution Network, which deconvolves the high-resolution heatmaps generated by HRNet to solve the scale variation challenge in bottom-up multi-person HPE. 

\textbf{Multi-task structures} are also employed in bottom-up HPE methods. Papandreou et al. \cite{papandreou2018personlab} introduced PersonLab to combine the pose estimation module and the person segmentation module for keypoints detection and association. PersonLab consists of short-range offsets (for refining heatmaps), mid-range offsets (for predicting the keypoints), and long-range offsets (for grouping keypoints into instances). Kocabas et al. \cite{kocabas2018multiposenet} presented a multi-task learning model with a pose residual net, named MultiPoseNet, which can perform keypoint prediction, human detection, and semantic segmentation tasks altogether. However, these methods are struggling in dealing with the variance of human scales, to address this problem, Luo et al. \cite{luo2021rethinking} introduced a method named SAHR (scale-adaptive heatmap regression) to optimize the joint standard deviation adaptively, which improved the tolerance of various human scales and labeling ambiguities in an efficient way.

\subsection{2D HPE Summary}\label{2DHPE Summary} 
In summary, the performance of 2D HPE has been significantly improved with the blooming of deep learning techniques. In recent years, deeper and more powerful networks have enhanced the performance of 2D single-person HPE methods such as DeepPose \cite{toshev2014deeppose} and Stacked Hourglass Network \cite{newell2016stacked}, as well as in 2D multi-person HPE like AlphaPose \cite{fang2017rmpe} and OpenPose \cite{cao2017realtime}.

Although promising performance has been achieved, there are several challenges in 2D HPE that need to be further addressed in future research. First is the reliable detection of individuals under significant occlusion \cite{chen2018cascaded}, e.g., in crowd scenarios. Person detectors in top-down 2D HPE methods may fail to identify the boundaries of largely overlapped human bodies. Similarly, the difficulty of keypoint association is more pronounced for bottom-up approaches in occluded scenes.
 
The second challenge is computation efficiency. Although some methods like OpenPose \cite{cao2017realtime} can achieve near real-time processing on special hardware with moderate computing power (e.g., 22 FPS with an Nvidia GTX 1080 Ti GPU), it is still  difficult to implement the networks on resource-constrained devices.  Real-world applications 
(e.g., gaming, AR, and VR) 
require more efficient HPE methods on commercial devices which can bring better interactive experiences for users.

Another challenge lies in the limited data for rare poses. Although the size of current datasets for 2D HPE is large enough (e.g., COCO dataset \cite{lin2014microsoft}) for normal pose estimation (e.g., standing, walking, running), these datasets have limited training data for unusual poses, e.g., falling. The data imbalance may cause model bias, resulting in poor performance on those poses. It would be useful to develop effective data generation or augmentation techniques to generate extra pose data for training more robust models. 


\section{3D human pose estimation}\label{3D}
3D HPE, which aims to predict the locations of body joints in 3D space, has attracted much interest in recent years since it can provide extensive 3D structure information related to the human body.
It can be applied to various applications (e.g., 3D movie and animation industries, virtual reality, and sports analysis). Although significant improvements have been achieved in 2D HPE, 3D HPE still remains a challenging task. 
Most existing works tackle 3D HPE from monocular images or videos, which is an ill-posed and inverse problem due to projection of 3D to 2D where one dimension is lost.
When multiple views are available or other sensors such as IMU and LiDAR are deployed, 3D HPE can be a well-posed problem employing information fusion techniques. Another limitation is that deep learning models are data-hungry and sensitive to the data collection environment. Unlike 2D HPE datasets where accurate 2D pose annotation can be easily obtained, collecting accurate 3D pose annotation is time-consuming and manual labeling is not practical. Also, datasets are usually collected from indoor environments with selected daily actions. Recent works \cite{Zhou_2017_ICCV,Yang_2018_CVPR,Wandt_2019_CVPR} revealed the poor generalization of models trained with biased datasets by cross-dataset inference. In this section, we first focus on 3D HPE from monocular RGB images and videos and then cover 3D HPE based on other sensors.

\subsection{3D HPE from monocular RGB images and videos}\label{3drgb}
The monocular camera is the most widely used sensor for HPE in both 2D and 3D scenarios. Recent progress in deep learning-based 2D HPE from monocular images and videos has enabled researchers to extend their works to 3D HPE. 

The reconstruction of 3D human poses from a single view of monocular images and videos is a nontrivial task that suffers from self-occlusions and other object occlusions, depth ambiguities, and insufficient training data. It is a severely ill-posed problem because different 3D human poses can be projected to a similar 2D pose projection.

The problem of occlusion can be alleviated by estimating 3D human pose from multi-view cameras. In a multi-view setting, the viewpoints association needs to be addressed. Thus deep learning-based 3D HPE methods are divided into three categories: single-view single-person 3D HPE, single-view multi-person 3D HPE, and multi-view 3D HPE. 

\subsubsection{Single-view single person 3D HPE} \label{3d1view1person}

Single-person 3D HPE approaches can be classified into skeleton-only and human mesh recovery (HMR) categories based on whether to reconstruct 3D human skeleton or to recover 3D human mesh by employing a human body model.

\noindent \textbf{A. Skeleton-only.} The skeleton-only methods estimate 3D human joints as the final output. They do not employ human body models to reconstruct 3D human mesh representation. These methods can be further divided into direct estimation approaches and 2D to 3D lifting approaches.



\begin{figure}[htp]
  \centering
  \includegraphics[width=0.8\linewidth]{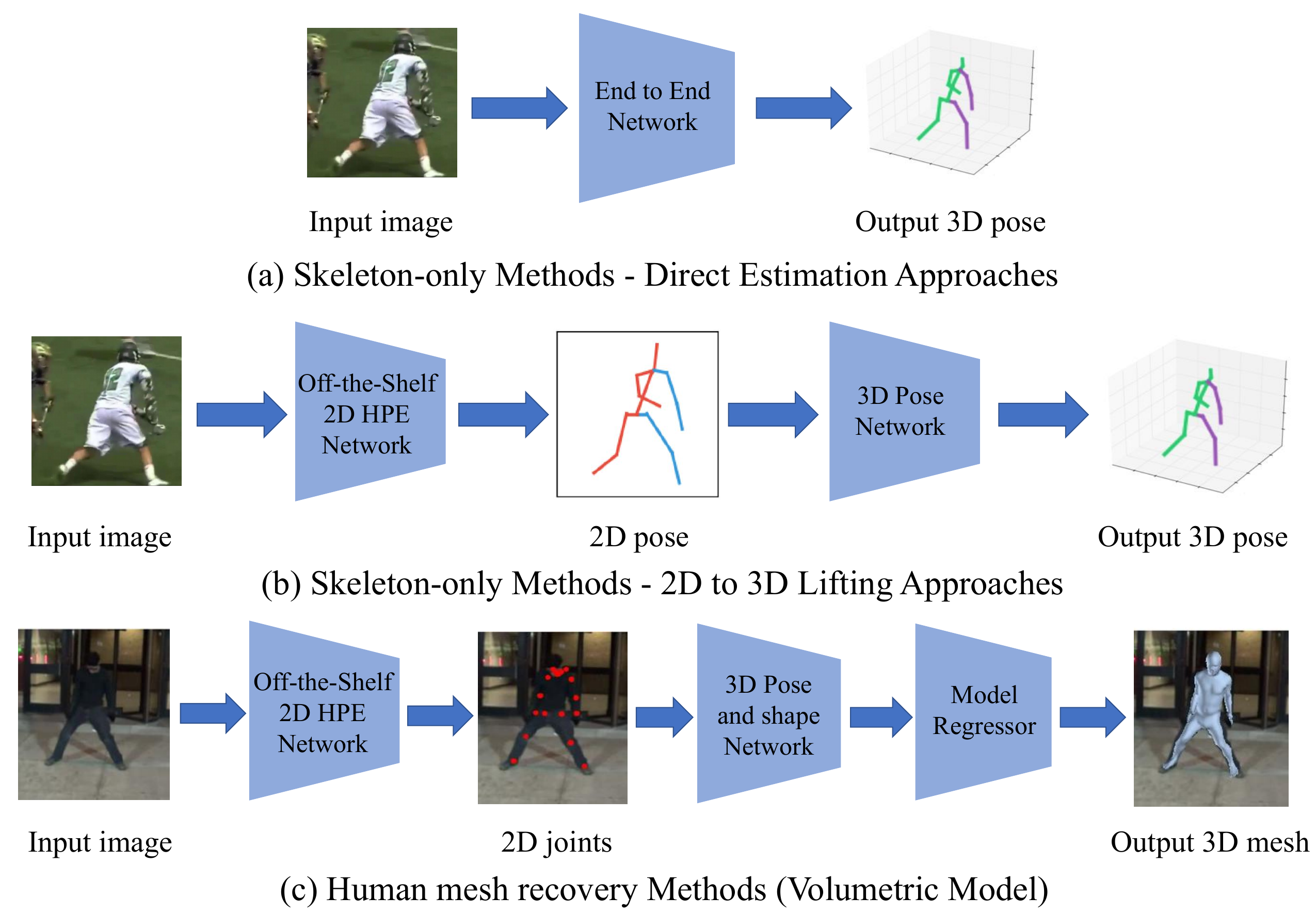}
\vspace{-0.3cm}
   \caption{\footnotesize{Single-person 3D HPE frameworks. (a) Direct estimation approaches directly estimate the 3D human pose from 2D images. (b) 2D to 3D lifting approaches leverage the predicted 2D human pose (intermediate representation) for 3D pose estimation. (c) Human mesh recovery methods incorporate parametric body models to recover a high-quality 3D human mesh. The 3D pose and shape parameters inferred by the 3D pose and shape network are fed into the model regressor to reconstruct 3D human mesh. Part of the figure is from \cite{Arnab_2019_CVPR}.}}
   \vspace{-5pt}
  \label{fig:3dsingleframeworks}
  \vspace{-5pt}
\end{figure}

\textbf{\textit{Direct estimation}:} As shown in Fig. \ref{fig:3dsingleframeworks}(a), direct estimation methods infer the 3D human pose from 2D images without intermediately estimating 2D pose representation, e.g., \cite{Li_2015_ICCV,Tekin2016,sun2017compositional,pavlakos2017volumetric,pavlakos2018ordinal}. Li and Chan \cite{Li_Chan_2014} employed a shallow network to train the body part detector with sliding windows and the pose coordinate regression synchronously. 
Sun et al. \cite{sun2017compositional} proposed a structure-aware regression approach. Instead of using a joint-based representation, they adopted a bone-based representation with more stability. A compositional loss was defined by exploiting the 3D bone structure with bone-based representation that encodes long-range interactions between the bones.
Pavlakos et al. \cite{pavlakos2017volumetric,pavlakos2018ordinal} introduced a volumetric representation to convert the highly non-linear 3D coordinate regression problem to a manageable form in a discretized space. The voxel likelihoods for each joint in the volume were predicted by a convolutional network. Ordinal depth relations of human joints were used to alleviate the need for accurate 3D ground truth poses. 


\textbf{\textit{2D to 3D lifting}:} Motivated by the recent success of 2D HPE, 2D to 3D lifting approaches that infer 3D human pose from the intermediately estimated 2D human pose have become a popular 3D HPE solution as illustrated in Fig. \ref{fig:3dsingleframeworks} (b). In the first stage, off-the-shelf 2D HPE models are employed to estimate 2D pose. Then 2D to 3D lifting is used to obtain 3D pose in the second stage, e.g., \cite{Chen20173DHP,martinez_2017_3dbaseline,Tekin2017fuse,zhou2019HEMlets,Moreno-Noguer2017,Li_2019_CVPR}. 
Benefiting from the excellent performance of state-of-the-art 2D pose detectors, 2D to 3D lifting approaches generally outperform direct estimation approaches. 
Martinez et al. \cite{martinez_2017_3dbaseline} proposed a fully connected residual network to regress 3D joint locations based on the 2D joint locations. Despite achieving state-of-the-art results at that time, the method could fail due to reconstruction ambiguity of over-reliance on the 2D pose detector. Tekin et al. \cite{Tekin2017fuse} and Zhou et al. \cite{zhou2019HEMlets} adopted 2D heatmaps instead of 2D pose as intermediate representations for estimating 3D pose. 
Wang et al. \cite{wang2018DRPose3D} developed a pairwise ranking CNN to predict the depth ranking of pairwise human joints. Then, a coarse-to-fine pose estimator was used to regress the 3D pose from 2D joints and the depth ranking matrix. Jahangiri and Yuille \cite{Jahangiri2017}, Sharma et al. \cite{Sharma_2019_ICCV}, and Li and Lee \cite{Li_2019_CVPR} first generated multiple diverse 3D pose hypotheses then applied ranking networks to select the best 3D pose.


Given that a human pose can be represented as a graph where the joints are the nodes and the bones are the edges, \textbf{Graph Convolutional Networks (GCNs)} have been applied to the 2D-to-3D pose lifting problem by showing promising performance \cite{Ci2019,Zhao_2019_Semantic_Graph,Choi_2020_ECCV_Pose2Mesh,Liu_2020_ECCV_weight_sharing,zeng2020srnet_ECCV}. 
Ci et al. \cite{Ci2019} proposed a Locally Connected Network (LCN), which leverages both a fully connected network and GCN to encode the relationship between local joint neighborhoods. LCN can overcome the limitations of GCN that the weight-sharing scheme harms the pose estimation model’s representation ability, and the structure matrix lacks the flexibility to support customized node dependence. Zhao et al. \cite{Zhao_2019_Semantic_Graph} also tackled the limitation of the shared weight matrix of convolution filters for all the nodes in GCN. A Semantic-GCN was proposed to investigate the semantic information and relationship, which is not explicitly represented in the graph. The semantic graph convolution (SemGConv) operation is used to learn channel-wise weights for edges. Both local and global relationships among nodes are captured since SemGConv and non-local layers are interleaved. Zhou et al. \cite{Zou_2021_mgcn} further introduced a novel modulated GCN network which consists of weight modulation and affinity modulation. The weight modulation exploits different modulation vectors for different nodes that disentangles the feature transformations. The affinity modulation explores additional joint correlations beyond the defined human skeleton. 

\textbf{The kinematic model} is an articulated body representation by connected bones and joints with kinematic constraints, which has gained increasing attention in 3D HPE in recent years. Many methods leverage prior knowledge based on the kinematic model such as skeletal joint connectivity information, joint rotation properties, and fixed bone-length ratios for plausible pose estimation, e.g., \cite{zhou2016deep,mehta2017vnect,Nie2017iccv,Wang_2019_ICCV,kundu2020kinematic,Xu_2020_CVPR,nie2020ECCV,Georgakis_ECCV2020_Hierarchical}. Zhou et al. \cite{zhou2016deep} embedded a kinematic model into a network as kinematic layers to enforce the orientation and rotation constraints. Nie et al. \cite{Nie2017iccv} and Lee et al. \cite{Lee_2018_ECCV} employed a skeleton-LSTM network to leverage joint relations and connectivity. Observing that human body parts have a distinct degree of freedom (DOF) based on the kinematic structure, Wang et al. \cite{Wang_2019_ICCV} and Nie et al. \cite{nie2020ECCV} proposed bidirectional networks to model the kinematic and geometric dependencies of the human skeleton. Kundu et al. \cite{kundu2020kinematic} \cite{Kundu_2020_CVPR} designed a kinematic structure preservation approach by inferring local-kinematic parameters with energy-based loss and explored 2D part segments based on the parent-relative local limb kinematic model. Xu et al. \cite{Xu_2020_CVPR} demonstrated that noise in the 2D joint is one of the key obstacles for accurate 3D pose estimation. Hence a 2D pose correction module was employed to refine unreliable 2D joints based on the kinematic structure. Zanfir et al. \cite{Zanfir_ECCV2020_Normalizing_Flows} introduced a kinematic latent normalizing flow representation (a sequence of invertible transformations applied to the original distribution) with differentiable semantic body part alignment loss functions.

3D HPE datasets are usually collected from controlled environments with selected common motions. It is difficult to obtain accurate 3D pose annotations for in-the-wild data. Thus 3D HPE for \textbf{in-the-wild data with unusual poses and occlusions} is still a challenge. 
To this end, a group of 2D to 3D lifting methods estimate the 3D human pose from in-the-wild images without 3D pose annotations such as \cite{Zhou_2017_ICCV,Habibie2019,Chen_2019_CVPR,Yang_2018_CVPR,Wandt_2019_CVPR}. Zhou et al. \cite{Zhou_2017_ICCV} proposed a weakly supervised transfer learning method that uses 2D annotations of in-the-wild images as weak labels. A 3D pose estimation module was connected with intermediate layers of the 2D pose estimation module. For in-the-wild images, 2D pose estimation module performed a supervised 2D heatmap regression and a 3D bone length constraint-induced loss was applied in the weakly supervised 3D pose estimation module. Habibie et al. \cite{Habibie2019} tailored a projection loss to refine the 3D human pose without 3D annotation. A 3D-2D projection module was designed to estimate the 2D body joint locations with the predicted 3D pose from the earlier network layer. The projection loss was used to update the 3D human pose without requiring 3D annotations. Inspired by \cite{Drover2018}, Chen et al. \cite{Chen_2019_CVPR} proposed an unsupervised lifting network based on the closure and invariance lifting properties with a geometric self-consistency loss for the lift-reproject-lift process. Closure means for a lifted 3D skeleton, after random rotation and re-projection, the resulting 2D skeleton will lie within the distribution of valid 2D poses. Invariance means when changing the viewpoint of 2D projection from a 3D skeleton, the re-lifted 3D skeleton should be the same.

Instead of estimating 3D human pose from monocular images, \textbf{videos can provide temporal information} to improve accuracy and robustness of 3D HPE, e.g., \cite{zhou2016sparseness,zhou2018monocap,Dabral_2018_ECCV,pavllo2019,Cheng2019occlusionaware,Cai2019Spatial-Temporal,wang2020motion,Tekin_2016_CVPR}. 
Hossain and Little \cite{Hossain_2018_ECCV} proposed a recurrent neural network using a Long Short-Term Memory (LSTM) unit with shortcut connections to exploit temporal information from sequences of human pose. Their method exploits the past events in a sequence-to-sequence network to predict temporally consistent 3D pose. Noticing that the complementary property between spatial constraints and temporal correlations is usually ignored by prior work, Dabral et al. \cite{Dabral_2018_ECCV}, Cai et al. \cite{Cai2019Spatial-Temporal}, and Li et al. \cite{Li2019boosting} exploited the spatial-temporal relationships and constraints (e.g., bone-length constraint and left-right symmetry constraint) to improve 3D HPE performance from sequential frames. Pavllo et al. \cite{pavllo2019} proposed a temporal convolution network to estimate 3D pose over 2D keypoints from consecutive 2D sequences. However, their method is based on the assumption that prediction errors are temporally non-continuous and independent, which may not hold in the presence of  occlusions \cite{Cheng2019occlusionaware}. Based on \cite{pavllo2019}, Chen et al. \cite{chen2020anatomy} added a bone direction module and bone length module to ensure human anatomy temporal consistency across video frames, while Liu et al. \cite{Liu_2020_CVPR} utilized the attention mechanism to recognize significant frames and model long-range dependencies in large temporal receptive fields. Zeng et al. \cite{zeng2020srnet_ECCV} employed the split-and-recombine strategy to address the rare and unseen pose problem. The human body is first split into local regions for processing through separate temporal convolutional network branches, then the low-dimensional global context obtained from each branch is combined for maintaining global coherence. 

\textbf{Transformer architectures} have become the model of choice in natural language processing due to the self-attention mechanism, and now are developing rapidly in the field of computer vision. Recent works have demonstrated the powerful global representation ability of transformer attention mechanism in various vision tasks \cite{khan2021transformers}. Zheng et al. \cite{zheng20213d} presented the first purely transformer-based approach, named PoseFormer, for 3D HPE without convolutional architectures involved. The spatial transformer module encodes local relationships between human body joints, and the temporal transformer module captures the global dependencies across frames in the entire sequence. 
Li et al. \cite{li2021mhformer} further designed a multi-hypothesis transformer to exploit spatial-temporal representations of multiple pose hypotheses. 
Zhao et al. \cite{poseformerv2} further proposed PoseFormerV2, which exploits a compact representation of lengthy skeleton sequences in the
frequency domain to efficiently scale up the receptive field and boost robustness to noisy 2D joint detection.

\noindent \textbf{B. Human Mesh Recovery (HMR).} HMR methods incorporate parametric body models such as SMPL \cite{SMPL:2015} to recovery human mesh as shown in Fig. \ref{fig:3dsingleframeworks}(c). The SMPL (Skinned Multi-Person Linear) model \cite{SMPL:2015} is a widely used model in 3D HPE, which can be modeled with natural pose-dependent deformations exhibiting soft-tissue dynamics. To learn how people deform with poses, there are 1786 high-resolution 3D scans of different subjects of poses with template mesh in SMPL to optimize the blend weights \cite{blendskinning}, pose-dependent blend shapes, the mean template shape, and the regressor from vertices to joint locations. The 3D pose can be obtained by using the model-defined joint regression matrix \cite{Kolotouros2019SPIN}.
There are also other popular volumetric models such as DYNA \cite{Dyna:SIGGRAPH:2015}, Stitched Puppet model \cite{Stitched_Puppet}, Frankenstein \& Adam \cite{Frankenstein_and_adam}, and GHUM \& GHUML(ite) \cite{GHUMGHUML}.  

\textbf{Volumetric models are used to recover high-quality human mesh}, providing extra shape information of the human body. As one of the most popular volumetric models, the SMPL model \cite{SMPL:2015} has been widely used in 3D HPE, e.g., 
\cite{Bogo:ECCV:2016,Kolotouros2019SPIN,zhu2019detailed,Arnab_2019_CVPR,Moon_I2L_MeshNet,Zhang_2020_CVPR_Object_Occluded,Li_2021_CVPR,pymaf2021}, because it is compatible with existing rendering engines. Tan et al. \cite{tan2017BMVC}, Tung et al. \cite{Tung2017}, Pavlakos et al. \cite{pavlakos2018humanshape}, and Omran et al. \cite{omran2018nbf} regressed SMPL parameters to reconstruct 3D human mesh. Instead of predicting SMPL parameters, Kolotouros et al. \cite{kolotouros2019cmr} regressed the locations of the SMPL mesh vertices using a Graph-CNN architecture. 
Kocabas et al. \cite{kocabas2020vibe} included the large-scale motion capture dataset AMASS \cite{AMASS2019} for adversarial training of their SMPL-based method named VIBE (Video Inference for Body Pose and Shape Estimation). VIBE leveraged AMASS to discriminate between real human motions and those predicted by the pose regression module. Since low-resolution visual content is more common in real-world scenarios than high-resolution visual content, existing well-trained models may fail when the resolution is degraded. Xu et al. \cite{xu_ECCV2020_contrastive_learning} introduced the contrastive learning scheme into a self-supervised resolution-aware SMPL-based network. The self-supervised contrastive learning scheme uses a self-supervision loss and a contrastive feature loss to enforce feature and scale consistency. Choi et al. \cite{TCMR_Choi_2021} presented a temporally consistent mesh recovery system (named TCMR) to smooth 3D human motion output using a bi-directional gated recurrent unit. Kolotouros et al. \cite{kolotouros2021prohmr} proposed a probabilistic model using conditional normalizing flow for 3D human mesh recovery from 2D evidence. Zheng et al. \cite{ce2021gtrs} designed a lightweight transformer-based method that can reconstruct human mesh from 2D human pose with a significant computation and memory cost reduction, while the performance is competitive with Pose2Mesh \cite{Choi_2020_ECCV_Pose2Mesh}. 

There are a few recent attempts to utilize \textbf{transformer} in HMR \cite{lin2021metro,ce2021gtrs,zheng2023potter,zheng2023feater}. Lin et al. proposed METRO \cite{lin2021metro} and MeshGraphormer \cite{lin2021_mesh_graphormer} that combine CNNs with transformer networks to regress SMPL mesh vertices from a single image. However, they pursued higher accuracy while sacrificing computation and memory. 
FeatER \cite{zheng2023feater} and POTTER \cite{zheng2023potter} reduced the computational and memory cost by proposing lightweight transformer architectures, while both of them outperform METRO by only requiring less than 10\%of total parameters and 15\% MACs.

There are several  \textbf{extended SMPL-based models} to address the limitations of the SMPL model such as high computational complexity, and lack of hands and facial landmarks. SMPLify \cite{Lassner2017} is an optimization method that fits the SMPL model to the detected 2D joints and minimizes the re-projection error. Pavlakos et al. \cite{Pavlakos2019SMPLX} introduced a new model, named SMPL-X, that can also predict fully articulated hands and facial landmarks. Following the SMPLify method, they also proposed SMPLify-X, which is an improved version learned from AMASS dataset \cite{AMASS2019}. Hassan et al. \cite{Hassan2019PROX} further extended SMPLify-X to PROX -- a method enforcing Proximal Relationships with Object eXclusion by adding 3D environmental constraints. Kolotouros et al. \cite{Kolotouros2019SPIN} integrated the regression-based and optimization-based SMPL parameter estimation methods to a new one named SPIN (SMPL oPtimization IN the loop) while employing  SMPLify in the training loop. The estimated 2D pose served as the initialization in an iterative optimization routine for producing a more accurate 3D pose and shape. 

Instead of using the SMPL-based model, {other models} have also been used for recovering 3D human pose or mesh, e.g.,\cite{Qammaz2019,Saito_2020_CVPR,Wang_ECCV2020_BLSM}. Chen et al. \cite{Cheng_2019_ICCV} introduced a Cylinder Man Model to generate occlusion labels for 3D data and performed data augmentation. A pose regularization term was introduced to penalize wrong estimated occlusion labels. Xiang et al. \cite{xiang2019monocular} utilized the Adam model \cite{Frankenstein_and_adam} to reconstruct the 3D motions. A 3D human representation, named 3D Part Orientation Fields (POFs), was introduced to encode the 3D orientation of human body parts in the 2D space. 
Wang et al. \cite{Wang_ECCV2020_BLSM} presented a new Bone-level Skinned Model of human mesh, which decouples bone modeling and identity-specific variations by setting bone lengths and joint angles. Fisch and Clark \cite{fisch2020orientation} introduced an orientation keypoints model which can compute full 3-axis joint rotations including yaw, pitch, and roll for 6D HPE. \\


\vspace{-0.5cm}
\subsubsection{Single-view multi-person 3D HPE}\label{3d1viewnperson}

\begin{figure}[htp]
  \centering
  \includegraphics[width=0.9\linewidth]{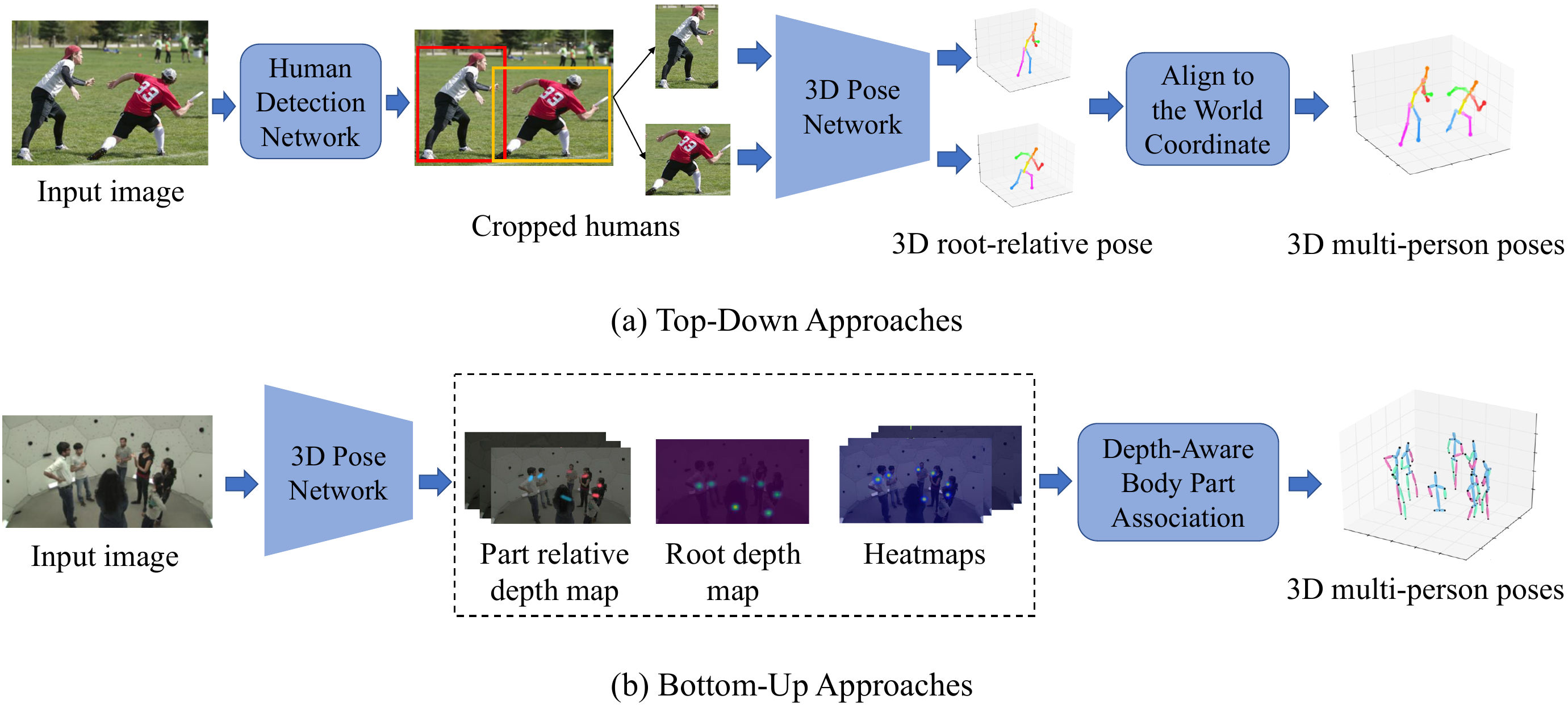}
\vspace{-0.3cm}
    \caption{\footnotesize{Illustration of the multi-person 3D HPE frameworks. (a) Top-Down methods first detect single-person regions by human detection network. For each single-person region, individual 3D poses can be estimated by 3D pose network. Then all 3D poses are aligned to the world coordinate. (b) Bottom-Up methods first estimate all body joints and depth maps, then associate body parts to each person according to the root depth and part relative depth. Part of the figure is from \cite{Zhen_ECCV2020_SMAP}.}}
  \label{fig:3dmultiframeworks}
  \vspace{-5pt}
\end{figure}

For 3D multi-person HPE from monocular RGB images or videos, similar categories as 2D multi-person HPE are noted here: top-down approaches and bottom-up approaches as shown in Fig. \ref{fig:3dmultiframeworks} (a) and Fig. \ref{fig:3dmultiframeworks} (b), respectively. The comparison between 2D top-down and bottom-up approaches in Section \ref{2D multi-person} is applicable to the 3D case. 

\textbf{Top-down approaches.}
Top-down approaches of 3D multi-person HPE first perform human detection to detect each individual  person. Then for each detected person, the absolute root (center joint of the human) coordinate and 3D root-relative pose are estimated by 3D pose networks. Based on the absolute root coordinate of each person and their root-relative pose, all poses are aligned to the world coordinate. Rogez et al. \cite{Rogez_LCRNet} localized candidate regions of each person to generate potential poses, and used a regressor to jointly refine the pose proposals. This localization-classification-regression method, named LCR-Net, performed well on the controlled environment datasets but could not generalize well to in-the-wild images. To address this issue, Rogez et al. \cite{Rogez_LCRNetpp} proposed LCR-Net++ by using synthetic data augmentation for the training data to improve performance. Zanfir et al. \cite{Zanfir2018} added semantic segmentation to the 3D multi-person HPE module with scene constraints. Additionally, the 3D temporal assignment problem was tackled by the Hungarian matching method for video-based multi-person 3D HPE. Moon et al. \cite{Moon_2019_ICCV_3DMPPE} introduced a camera distance-aware approach under top-down pipeline. The RootNet estimated the camera-centered coordinates of human body's roots. 
Then the root-relative 3D pose of each cropped human was estimated by the proposed PoseNet. Benzine et al. \cite{Benzine_2020_CVPR} proposed a single-shot approach named PandaNet (Pose estimAtioN and Detection Anchor-based Network). A low-resolution anchor-based representation was introduced to avoid the occlusion problem. A pose-aware anchor selection module was developed to address the overlapping problem by removing ambiguous anchors. An automatic weighting of losses associated with different scales was used to handle the imbalance issue of different sizes of people. Li et al. \cite{Li_ECCV2020_HMOR} tackled the  lack of  global information in the top-down approaches. They adopted a Hierarchical Multi-person Ordinal Relations method to leverage body-level semantic and global consistency for encoding the interaction information hierarchically.

\textbf{Bottom-up approaches.} In contrast to top-down approaches, bottom-up approaches first produce all body joint locations and depth maps, then associate body parts to each person according to the root depth and part relative depth. A key challenge of bottom-up approaches is how to group human body joints belonging to each person. Zanfir et al. \cite{ZanfirNIPS2018} formulated the person grouping problem as a binary integer programming (BIP) problem. A limb scoring module was used to estimate candidate kinematic connections of detected joints and a skeleton grouping module assembled limbs into skeletons by solving the BIP problem. Nie et al. \cite{Nie_2019_ICCV} proposed a Single-stage multi-person Pose Machine (SPM) to define the unique identity root joint for each person. The body joints were aligned to each root joint by using dense displacement maps. However, this method is limited in that only paired 2D images and 3D pose annotations can be used for supervised learning. Without paired 2D images and 3D pose annotations, Kundu et al. \cite{Kundu_ECCV2020_CrossModal} proposed a frozen network to exploit the shared latent space between two diverse modalities under a practical deployment paradigm such that  the learning could be cast as a cross-model alignment problem. Fabbri et al. \cite{fabbri2020compressed} developed a distance-based heuristic for linking joints in a multi-person setting. Specifically, starting from detected
heads (i.e., the joint with the highest confidence), the remaining joints are connected by selecting the closest ones in terms of 3D Euclidean distance. Chen et al. \cite{Cheng_2021_TDBU} integrated top-down and bottom-up approaches in their method. A top-down network first estimates joint heatmaps inside each bounding box, then a bottom-up network incorporates estimated joint heatmaps to handle the scale variation. 

Another challenge of bottom-up approaches is occlusion. To cope with this challenge, Metha et al. \cite{Mehta20183DV} developed an Occlusion-Robust Pose-Maps (ORPM) approach to incorporate redundancy into the location-maps formulation, which facilitates person association in the heatmaps, especially for occluded scenes. Zhen et al. \cite{Zhen_ECCV2020_SMAP} leveraged a depth-aware part association algorithm to assign joints to individuals by reasoning about inter-person occlusion and bone-length constraints. Mehta et al. \cite{mehtaxnect} quickly inferred intermediate 3D pose of visible body joints regardless of the accuracy. Then the completed 3D pose is reconstructed by inferring occluded joints using learned pose priors and global context. The final 3D pose was refined by applying temporal coherence and fitting the kinematic skeletal model.

\textbf{Comparison of top-down and bottom-up approaches.}
Top-down approaches usually achieve promising results by relying on state-of-the-art person detection methods and single-person HPE methods. But the computational complexity and the inference time may become excessive with the increase in the number of humans, especially in crowded scenes. Moreover, since top-down approaches first detect the bounding box for each person, global information in the scene may get neglected. The estimated depth of cropped region may be inconsistent with the actual depth ordering and the predicted human bodies may be placed in overlapping positions. On the contrary, the bottom-up approaches enjoy linear computation and time complexity. However, 
if the goal is to recover 3D body mesh, it is not straightforward for the bottom-up approaches to reconstruct human body meshes. For top-down approaches, after detecting each individual person, human body mesh of each person can be easily recovered by incorporating the 3D single-person human mesh recovery method. 
While for the bottom-up approaches, an additional model regressor module is needed to reconstruct human body meshes based on the final 3D poses.

\subsubsection{Multi-view 3D HPE} \label{3dmultiview}

Partial occlusion is a challenging problem for 3D HPE in the single-view setting. The natural solution to overcome this problem is to estimate 3D human pose from multiple views, since the occluded part in one view may become visible in other views. In order to reconstruct the 3D pose from multiple views, the association of corresponding locations between different cameras needs to be resolved. We do not specify single-person or multi-person in this category since the multi-view setting is deployed mainly for multi-person pose estimation. 

A group of methods \cite{Qiu_2019_ICCV,Liang_2019_ICCV,Dong_2019_CVPR,Tu_ECCV2020_VoxelPose,dong2021shape}  used body models to tackle the association problem by optimizing model parameters to match the model projection with the 2D pose. The widely used 3D pictorial structure model \cite{3DPictorial} is such a model. However, these methods usually need large memory and expensive computational cost, especially for multi-person 3D HPE under multi-view settings. Rhodin et al. \cite{Rhodin_2018_CVPR} employed a multi-view consistency constraint in the network, however it requires a large amount of 3D ground-truth training data. To overcome this limitation, Rhodin et al. \cite{Rhodin_2018_ECCV} further proposed an encoder-decoder framework to learn the geometry-aware 3D latent representation from multi-view images and background segmentation without 3D annotations. 
Chen et al. \cite{chen_Crowded_ECCV2020}, Mitra et al. \cite{Mitra_2020_CVPR}, Zhang et al.\cite{20204DAssociation}, and Huang et al. \cite{Huang_ECCV2020_DynamicMatching}
proposed multi-view matching frameworks to reconstruct 3D human pose across all viewpoints with consistency constraints. Pavlakos et al. \cite{Pavlakos_mutiview_2017_CVPR} and Zhang et al. \cite{zhang2020adafuse} aggregated the 2D keypoint heatmaps of multi-view images into a 3D pictorial structure model based on all the calibrated camera parameters. However, when multi-view camera environments change, the model needs to be retrained. 
Qiu et al. \cite{Qiu_2019_ICCV}, and Kocabas et al. \cite{Kocabas_2019_CVPR} employed epipolar geometry to match paired multi-view poses for 3D pose reconstruction and generalized their methods to new multi-view camera environments. It should be noted that matching each pair of views separately without the cycle consistency constraint may lead to incorrect 3D pose reconstructions \cite{Dong_2019_CVPR}. Tu et al. \cite{Tu_ECCV2020_VoxelPose} aggregated all the features in each camera view in the 3D voxel space to avoid incorrect estimation in each camera view. A cuboid proposal network and a pose regression network were designed to localize all people and to estimate the 3D pose, respectively. When given sufficient viewpoints (more than ten), it is not practical to use all viewpoints for 3D pose estimation. Pirinen et al. \cite{Pirinen_NIPS2019} proposed a self-supervised reinforcement learning approach to select a small set of viewpoints to reconstruct the 3D pose via triangulation. Wang et al. \cite{wang2021mvp} introduced a transformer-based model that directly regresses 3D poses from multi-view images without relying on any intermediate task. The proposed Multi-view Pose transformer (MvP) was designed to represent query embedding of multi-person joints. The multi-view information was fused by a novel geometrically guided attention mechanism.  

Besides accuracy, \textbf{lightweight architecture, fast inference time, and efficient adaptation} to new camera settings also need to be taken into consideration in multi-view HPE. In contrast to \cite{Dong_2019_CVPR} which matched all view inputs together, Chen et al. \cite{Chen_2020_CVPR} applied an iterative processing strategy to match 2D poses of each view with the 3D pose while the 3D pose was updated iteratively. Compared to previous methods whose running time may explode with  the increase in the  number of cameras, their method  has linear time complexity. 
Remelli et al. \cite{Remelli_2020_CVPR} encoded images of each view into a unified latent representation so that the feature maps were disentangled from camera viewpoints. As a lightweight canonical fusion, these 2D representations are lifted to the 3D pose using a GPU-based Direct Linear Transform to accelerate processing. In order to improve the generalization ability of the multi-view fusion scheme, Xie et al. \cite{Xie_2020_CVPR} proposed a pre-trained multi-view fusion model (MetaFuse), which can be efficiently adapted to new camera settings with few labeled data. They deployed the model-agnostic meta-learning framework to learn the optimal initialization of the generic fusion model for adaptation. To reduce the computational cost of the VoxelPose \cite{Tu_ECCV2020_VoxelPose}, Ye et al. presented Faster VoxelPose \cite{ye2022faster} which re-projects the feature volume to three two-dimensional coordinate planes for estimating X, Y, Z coordinates from them separately. The fps of Faster VoxelPose is 31.1, which is almost 10 $\times$ speed up compared to VoxelPose.

\subsection{3D HPE from other sources}\label{3dother}
Although a monocular RGB camera is the most common device used for 3D HPE, other sensors (e.g., depth sensor, IMUs, and radio frequency device) are also used for this purpose.

\textbf{Depth and point cloud sensors}: Depth sensors have gained more attention recently for 3D computer vision tasks due to their low-cost and increased utilization. As one of the key challenges in 3D HPE, the depth ambiguity problem can be alleviated by using depth sensors. Yu et al. \cite{Yu_2019_DoubleFusion}, Xiong et al. \cite{Xiong_2019_A2J}, Kadkhodamohammadi et al. \cite{Kadkhodamohammadi2017}, and Zhi et al. \cite{Zhi_ECCV2020_TexMesh} utilized depth images to estimate 3D human pose. 

Compared with depth images, point clouds can provide more information. The state-of-the-art point cloud feature extraction techniques, PointNet \cite{qi2017pointnet} and PointNet++ \cite{qi2017pointnet++}, have demonstrated excellent performance for classification and segmentation tasks. Jiang et al. \cite{Jiang_2019_ICCV} and Wang et al. \cite{Wang_2020_CVPR} combined PointNet++ with the 3D human body model to recover 3D human mesh. 

\textbf{IMUs with monocular images}: Wearable Inertial Measurement Units (IMUs) can track the orientation and acceleration of human body parts by recording  motions without object occlusions and clothes obstructions. Marcard et al.\cite{von2017sparse,Marcard_2018_ECCV}, Huang et al. \cite{Huang_2018_DIP}, Zhang et al. \cite{Zhang_2020_CVPR}, and Huang et al. \cite{Huang_2020_WACV} proposed IMU-based HPE methods to reconstruct 3D human pose. However, drifting may occur over time when using IMUs. 

\textbf{Radio frequency device}: Radio Frequency (RF) based sensing technology has also been used for 3D HPE, e.g., \cite{Zhao_2018_RF3D} and \cite{Zhao_2019_RF}. The ability to traverse walls and bounce off human bodies in the WiFi range without carrying wireless transmitters is the major advantage of deploying an RF-based sensing system. Also, privacy can be preserved due to non-visual data. However, RF signals have a relatively low spatial resolution compared to visual camera images, and RF systems have been shown to generate coarse 3D pose estimation. 

\textbf{Other sensors/sources:} Besides using the aforementioned sensors, Isogawa et al. \cite{Isogawa_2020_CVPR} estimated 3D human pose from the 3D spatio-temporal histogram of photons captured by a non-line-of-sight (NLOS) imaging system. Some methods \cite{tome2019xr,tome2020selfpose,xu2019mo}  tackled the egocentric 3D pose estimation via a fish-eye camera. Saini et al. \cite{Nitin_ICCV_19} estimated human motion using images captured by multiple Autonomous Micro Aerial Vehicles (MAVs). Clever et al. \cite{Clever_2020_CVPR} focused on the HPE of the rest position in bed from pressure images which were collected by a pressure sensing mat.   

\subsection{3D HPE Summary}\label{3DHPE_Summary}
3D HPE has made significant advancements in recent years. 
Since a large number of 3D HPE methods apply the 2D to 3D lifting strategy, 
the performance of 3D HPE has been improved considerably due to the progress made in 2D HPE. Some 2D HPE methods such as OpenPose \cite{cao2017realtime}, AlphaPose \cite{fang2017rmpe}, and HRNet \cite{sun2019deep} have been extensively used as 2D pose detectors in 3D HPE methods. Besides the 3D pose, some methods also recover 3D human mesh from images or videos, e.g., \cite{Kolotouros2019SPIN,kocabas2020vibe,Zeng_cvpr2020,zhou2020Disentanglement}. However, despite the progress made so far, there are still several challenges. 

One challenge is model generalization. High-quality 3D ground truth pose annotations 
depend on motion capture systems that cannot be easily deployed in a random environment. 
Therefore, the existing datasets are mainly captured in constrained scenes.
The state-of-the-art methods can achieve promising results on these datasets, but their performance degrades when applied to in-the-wild data. 
It is possible to leverage gaming engines to generate synthetic datasets with diverse poses and complex scenes, e.g., SURREAL dataset \cite{SURREAL2016} and GTA-IM dataset \cite{GTA-IM-ECCV2020}. However, learning from synthetic data may not achieve the
desired performance due to a gap between synthetic and real data distributions.

Like 2D HPE, robustness to occlusion and computational efficiency are two key challenges for 3D HPE as well. 
The performance of current 3D HPE methods drops considerably in crowded scenarios due to severe mutual occlusion and possibly low-resolution content of each person. 3D HPE is more computation-demanding than 2D HPE. For example, 2D to 3D lifting approaches rely on 2D poses as intermediate representations for inferring 3D poses. It is critical to develop computationally efficient 2D HPE pipelines while maintaining high accuracy for pose estimation. 

\section{Datasets and Evaluation Metrics}\label{dataset}
Datasets are very much needed in conducting HPE. They are also necessary to provide a fair comparison among different algorithms. 
In this section, we present the most widely used datasets and evaluation metrics for evaluating 2D and 3D deep learning-based HPE methods. 
The results achieved by existing methods on these popular datasets are summarized. 

\subsection{Datasets for 2D HPE}
Although there are several datasets used for 2D HPE tasks before 2014, only a few recent works use these datasets because they have several shortcomings such as a lack of diverse object movements and limited data. Since deep learning-based approaches are fueled by a large amount of training data, this section mainly discusses the recent and large-scale 2D human pose datasets. 



\href{http://human-pose.mpi-inf.mpg.de/#}{\textbf{Max Planck Institute for Informatics (MPII) Human Pose Dataset}} \cite{andriluka20142d} is a popular dataset for evaluation of articulated HPE 
which includes around 25,000 images containing over 40,000 individuals with annotated body joints. 
Rich annotations including body part occlusions, 3D torso, and head orientations are labeled by workers on Amazon Mechanical Turk. Images in MPII are suitable for 2D single-person or multi-person HPE. 

\href{https://cocodataset.org/#home}{\textbf{Microsoft Common Objects in Context (COCO) Dataset}} \cite{lin2014microsoft} is the most widely used large-scale dataset. It has more than 330,000 images and 200,000 labeled subjects with keypoints, and each individual person is labeled with 17 joints.
In addition, Jin et al. \cite{jin2020whole} proposed COCO-WholeBody Dataset with whole-body annotations for HPE.

\href{https://posetrack.net/}{\textbf{PoseTrack Dataset}} \cite{PoseTrack} is a large-scale dataset for HPE and articulated tracking in videos, including body part occlusion and truncation in crowded environments. There are two versions for PoseTrack dataset: PoseTrack2017 \cite{PoseTrack} contains 514 video sequences with 16,219 pose annotations, which are split into 250 (training), 50 (validation), and 214 (testing) sequences. PoseTrack2018 \cite{andriluka2018posetrack} has 1,138 video sequences with 153,615 pose annotations, which are divided into 593 for training, 170 for validation, and 375 for testing. Each person in PoseTrack is labeled with 15 joints and an additional label for keypoint visibility.

We refer the readers to the original references for details about other datasets including \textbf{LSP} \cite{johnson2010clustered}, \textbf{FLIC} \cite{sapp2013modec}, \textbf{AIC-HKD} \cite{wu2017ai}, \textbf{CrowdPose} \cite{li2019crowdpose}, \textbf{Penn Action} \cite{zhang2013actemes}, \textbf{J-HMDB} \cite{Jhuang:ICCV:2013}, and \textbf{HiEve} \cite{lin2020human}. A summary of these datasets is presented in Table \ref{tab: 2Ddatasets}. 

\begin{table}[]
\centering
\renewcommand\arraystretch{0.9}
\setlength\tabcolsep{25.0pt} 
  \caption{\footnotesize{Datasets for 2D HPE.} }
  \vspace{-5pt}
  \resizebox{\linewidth}{!}{\begin{tabular}{|c|c|c|c|ccc|c|}
\hline
\multicolumn{8}{|c|}{\textbf{Image-based datasets} }                                                                                                                                                                                                                                \\ \hline
\multirow{2}{*}{Name} & \multirow{2}{*}{Year} & \multirow{2}{*}{\begin{tabular}[c]{@{}c@{}}Single-Person\\ /Multi-Person\end{tabular}} & \multirow{2}{*}{Joints} & \multicolumn{3}{c|}{Number of images } & \multirow{2}{*}{\begin{tabular}[c]{@{}c@{}}Evaluation \\ protocol\end{tabular}} \\ \cline{5-7}
                      &                       &                                                                                        &                         & Train   & Val    & Test   &                                                                                 \\ \hline
LSP \cite{johnson2010clustered}                  & 2010                  & Single                                                                                 & 14                      & 1k      & -      & 1k     & PCP                                                                             \\ \hline
LSP-extended \cite{johnson2011learning}          & 2011                  & Single                                                                                 & 14                      & 10k     & -      & -      & PCP                                                                             \\ \hline
FLIC \cite{sapp2013modec}                  & 2013                  & Single                                                                                 & 10                      & 5k      & -      & 1k     & PCP\&PCK                                                                        \\ \hline
FLIC-full \cite{sapp2013modec}              & 2013                  & Single                                                                                 & 10                      & 20k     & -      & -      & PCP\&PCK                                                                        \\ \hline
FLIC-plus \cite{tompson2014joint}             & 2014                  & Single                                                                                 & 10                      & 17k     & -      & -      & PCP\&PCK                                                                        \\ \hline
\multirow{2}{*}{MPII \cite{andriluka20142d}} & \multirow{2}{*}{2014} & Single                                                                                 & 16                      & 29k     & -      & 12k    & PCPm/PCKh                                                                       \\
                      &                       & Multiple                                                                               & 16                      & 3.8k    & -      & 1.7k   & mAp                                                                             \\ \hline
COCO2016 \cite{lin2014microsoft}              & 2016                  & Multiple                                                                               & 17                      & 45k     & 22k    & 80k    & AP                                                                              \\ \hline
COCO2017 \cite{lin2014microsoft}              & 2017                  & Multiple                                                                               & 17                      & 64k     & 2.7k   & 40k    & AP                                                                              \\ \hline
AIC-HKD \cite{wu2017ai}               & 2017                  & Multiple                                                                               & 14                      & 210k    & 30k    & 60k    & AP                                                                              \\ \hline
CrowdPose \cite{li2019crowdpose}               & 2019                  & Multiple                                                                               & 14                      & 10k    & 2k    & 8k    & mAP                                                                              \\ \hline
\multicolumn{8}{|c|}{\textbf{Video-based   datasets} }                                                                                                                                                                                                                                \\ \hline
\multirow{2}{*}{Name} & \multirow{2}{*}{Year} & \multirow{2}{*}{\begin{tabular}[c]{@{}c@{}}Single-Person\\ /Multi-Person\end{tabular}} & \multirow{2}{*}{Joints} & \multicolumn{3}{c|}{Number of videos} & \multirow{2}{*}{\begin{tabular}[c]{@{}c@{}}Evaluation \\ protocol\end{tabular}} \\ \cline{5-7}
                      &                       &                                                                                        &                         & Train   & Val    & Test   &                                                                                 \\ \hline
Penn Action \cite{zhang2013actemes}           & 2013                  & Single                                                                                 & 13                      & 1k      & -      & 1k     & -                                                                               \\ \hline
J-HMDB \cite{Jhuang:ICCV:2013}                & 2013                  & Single                                                                                 & 15                      & 0.6k    & -      & 0.3k   & -                                                                               \\ \hline
PoseTrack2017 \cite{PoseTrack}             & 2017                  & Multiple                                                                               & 15                      & 250     & 50     & 214    & mAP                                                                             \\ \hline
PoseTrack2018 \cite{andriluka2018posetrack}             & 2018                  & Multiple                                                                               & 15                      & 593     & 170     & 375    & mAP                                                                             \\ \hline
HiEve \cite{lin2020human}             & 2020                  & Multiple                                                                               & 14                      & 19     & -     & 13    & mAP                                                                             \\ \hline
\end{tabular}
}
\label{tab: 2Ddatasets}
\vspace{-15pt}
\end{table}

\subsection{Evaluation Metrics for 2D HPE}
It is difficult to precisely evaluate the performance of HPE because there are many features and requirements that need to be considered (e.g., upper/full human body, single/multiple pose estimation, the size of human body). As a result, many evaluation metrics have been used for 2D HPE. Here we summarize the commonly used ones.

\textbf{Percentage of Correct Parts (PCP)} \cite{eichner20122d} is a measure commonly used in early works on 2D HPE, which evaluates stick predictions to report the localization accuracy for limbs. The localization of limbs is determined when the distance between the predicted joint and ground truth joint is less than a fraction of the limb length (between 0.1 to 0.5). In some works, the PCP measure is also referred to as PCP@0.5, where the threshold is 0.5. This measure is used for single-person HPE evaluation. However, PCP has not been widely implemented in the latest works because it penalizes the limbs with short lengths which are hard to detect. The performance of a model is considered better when it has a higher PCP measure. In order to address the drawbacks of PCP, Percentage of Detected Joints (PDJ) is introduced, where a predicted joint is considered as detected if the distance between predicted joints and true joints is within a certain fraction of the torso diameter\cite{toshev2014deeppose}.

\textbf{Percentage of Correct Keypoints (PCK)} \cite{yang2012articulated} is also used to measure the accuracy of localization of different keypoints within a given threshold. The threshold is set to 50 percent of the head segment length of each test image and it is denoted as PCKh@0.5. PCK is referred to as PCK@0.2 when the distance between detected joints and true joints is less than 0.2 times the torso diameter. The higher the PCK value, the better model performance is regarded.

\textbf{Average Precision (AP) and Average Recall (AR)}. AP measure is an index to measure the accuracy of keypoints detection according to precision (the ratio of true positive results to the total positive results) and recall (the ratio of true positive results to the total number of ground truth positives). AP computes the average precision value for recall over 0 to 1. AP has several similar variants. For example, Average Precision of Keypoints (APK) is introduced in \cite{yang2012articulated}. Mean Average Precision (mAP), which is the mean of average precision over all classes, is a widely used metric on the MPII and PoseTrack datasets. Average Recall (AR) is another metric used in the COCO keypoint evaluation \cite{lin2014microsoft}. Object Keypoint Similarity (OKS) plays a similar role as the Intersection over Union (IoU) in object detection and is used for AP or AR. This measure is computed from the scale of the subject and the distance between predicted points and ground truth points. The COCO evaluation usually uses mAP across 10 OKS thresholds as the evaluation metric.

\begin{table}[!htbp]
\renewcommand\arraystretch{0.9}
\centering
\vspace{-5pt}
\setlength\tabcolsep{25.0pt} 
  \caption{\footnotesize{Comparison of different methods on the MPII dataset for 2D single-person HPE using PCKh@0.5 measure (i.e., the threshold is equal to 50 percent of the head segment length of each test image). \textbf{H}: Heatmap; \textbf{R}: Regression. (\textcolor{red}{Red}: best; \textcolor{blue}{Blue}: second best)
  }}
  \vspace{-8pt}
  \resizebox{\linewidth}{!}{
\begin{tabular}{|c|c|c|c|c|c|c|c|c|c|c|c|c|}
\hline
\multicolumn{13}{|c|}{Max Planck Institute for Informatics (MPII)}  \\ \hline
\multirow{-1}{*}{}  & \multirow{-1}{*}{Year} & \multirow{-1}{*}{Method} &\multirow{-1}{*}{Head} &\multirow{-1}{*}{Shoulder} & \multirow{-1}{*}{Elbow} & \multirow{-1}{*}{Wrist} & \multirow{-1}{*}{Hip} & \multirow{-1}{*}{Knee} & \multirow{-1}{*}{Ankle} & \multirow{-1}{*}{Mean} & \multirow{-1}{*}{Params(M)} & \multirow{-1}{*}{FLOPs(G)}\\ \hline   

&2017	&\cite{chu2017multi}	&\color{blue}98.5	&96.3	&91.9	&88.1	&90.6	&88.0 	&85.0	    &91.5 &- &-\\ 
&2017	&\cite{yang2017learning}	&\color{blue}98.5	&96.7	&92.5	&88.7	&91.1	&88.6	&86.0	    &92.0 &7.3 &14.7\\ 
&2018	&\cite{ke2018multi}	&\color{blue}98.5	&96.8	&92.7	&88.4	&90.6	&89.3	&86.3	&92.1 &- &-\\ 
&2018	&\cite{tang2018deeply}	&98.4	&96.9	&92.6	&88.7	&\color{blue}91.8	&89.4	&86.2	&92.3 &15.5 &33.6\\ 
&2018	&\cite{Xiao_2018_ECCV}	&\color{blue}98.5	&96.6	&91.9	&87.6	&91.1	&88.1	&84.1	&91.5 &- &-\\ 
&2019	&\cite{sun2019deep}	&\color{red}98.6	&96.9	&92.8	&89.0 	&91.5	&89.0 	&85.7	&92.3 &28.5 &9.5\\ 
&2019	&\cite{zhang2019human}	&\color{red}98.6	&97.0 	&92.8	&88.8	&91.7	&89.8	&\color{blue}86.6	&92.5 &23.9 &41.4\\ 
&2019	&\cite{li2019rethinking}	&98.4	&\color{blue}97.1	&\color{blue}93.2	&\color{blue}89.2	&\color{red}92.0 	&\color{blue}90.1	&85.5	&92.6 &- &- \\ 
&2020	&\cite{artacho2020unipose}	&-	    &-	    &-	    &-	    &-	    &-	    &-	    &\color{blue}92.7 &- &- \\ 
&2020	&\cite{cai2020learning}	&\color{blue}98.5	&\color{red}97.3	&\color{red}93.9	&\color{red}89.9	&\color{red}92.0 	&\color{red}90.6	&\color{red}86.8	&\color{red}93.0 &- &- \\
&2021	&\cite{Li_2021_ICCV}	&97.1	&95.9	&90.4	&86.0	&89.3 	&87.1	&82.5	&90.2 &28.1 &- \\
\multirow{-17}{*}{H}&2022	&\cite{li2022simcc}	&97.2	&96.0	&90.4	&85.6	&89.5 	&85.8	&81.8	&90.0 &- &-\\ \hline
&2017	&\cite{sun2017compositional}	&97.5	&94.3	&87.0	&81.2	&86.5	&78.5	&75.4	&86.4 &- &-\\
&2019   &\cite{zhang2019fast}  &98.3 &96.4 &91.5 &87.4 &90.0 &87.1 &83.7 &91.1 &3.0 &9.0\\ 
&2019	&\cite{luvizon2019human}	&98.1	&96.6	&92.0	&87.5	&90.6	&88	&82.7	&91.2 &- &-\\
&2021	&\cite{li2021pose}	&97.3	&96.0	&90.6	&84.5	&89.7	&85.5	&79.0	&89.5 &- &- \\
&2021	&\cite{mao2021tfpose}	&98.0	&95.9	&91.0	&86.0	&89.8	&86.6	&82.6	&90.4 &- &-\\
\multirow{-8}{*}{R} &2022	&\cite{mao2022poseur}	&-	&-	&-	&-	&-	&-	&-	&90.5 &- &-
 \\\hline
\end{tabular}
}
    \begin{tablenotes}
      \tiny
      \item Note: \cite{insafutdinov2016deepercut}, \cite{sun2019deep}, \cite{li2019rethinking}, \cite{cai2020learning}, \cite{Rogez_LCRNet},\cite{Li_2021_ICCV} are 2D multi-person HPE methods, which are also applied to the single-person case here.
    \end{tablenotes}
\label{MPII dataset on testing, using PCKh@0.5}
\vspace{-10pt}
\end{table}

\begin{table}[!htbp]
\centering
\setlength\tabcolsep{25.0pt} 
  \caption{\footnotesize{Comparison of different 2D multi-person HPE methods on the test-dev set of the COCO dataset using AP measure (AP.5: AP at OKS = 0.50; AP.75: AP at OKS = 0.75; AP(M) is used for medium objects; AP(L) is used for large objects). \textbf{Extra}: extra data is used for training. \textbf{T}: Top-down; \textbf{B}: Bottom-up
  }}
  \vspace{-5pt}
  \resizebox{\linewidth}{!}{
\begin{tabular}{|c|c|c|c|c|c|c|c|c|c|c|c|c|}
\hline
\multicolumn{13}{|c|}{Microsoft  Common  Objects  in  Context  (COCO)}  \\ \hline
\multirow{-1}{*}{}  & \multirow{-1}{*}{Year} & \multirow{-1}{*}{Method} &\multirow{-1}{*}{\bf Extra} &\multirow{-1}{*}{Backbone} & \multirow{-1}{*}{Input size} & \multirow{-1}{*}{AP} & \multirow{-1}{*}{AP.5} & \multirow{-1}{*}{AP.75} & \multirow{-1}{*}{AP(M)} & \multirow{-1}{*}{AP(L)} & \multirow{-1}{*}{Params(M)} & \multirow{-1}{*}{FLOPs(G)}\\ \hline
&2019	&\cite{sun2019deep}	&no	&HRNet-W32	        &384$\times$288	&74.9	&92.5	&82.8	&71.3	&80.9 &28.5 &16.0\\
&2019	&\cite{sun2019deep}	&no	&HRNet-W48	        &384$\times$288	&75.5	&92.5	&83.3	&71.9	&81.5 &63.6 &32.9\\
&2019	&\cite{sun2019deep}	&yes	&HRNet-W48	        &384$\times$288	&77.0 	&92.7	&84.5	&73.4	&83.1 &63.6 &32.9\\
&2019	&\cite{li2019rethinking}	&yes	&4xResNet-50	    &384$\times$288	&77.1	&\color{blue}93.8	&84.6	&73.4	&82.3 &- &-\\
&2020	&\cite{zhang2020distribution}	&no	&HRNet-W48 	        &384$\times$288	&76.2	&92.5	&83.6	&72.5	&82.4 &63.6 &32.9\\
&2020	&\cite{zhang2020distribution}	&yes	&HRNet-W48	        &384$\times$288	&77.4	&92.6	&84.6	&73.6	&\color{blue}83.7 &63.6 &32.9 \\
&2020	&\cite{cai2020learning}	&no	&4xRSN-50	        &384$\times$288	&\color{blue}78.6	&\color{red}94.3	&\color{red}86.6	&\color{blue}75.5	&83.3 &111.8 &65.9\\
&2021	&\cite{yang2021transpose}	&no	&HRNet-W48	        &256$\times$192	&75.0	&92.2	&82.3	&71.3	&81.1 &17.5 &21.8\\
&2021	&\cite{Li_2021_ICCV}	&no	 &HRNet-W48	        &384$\times$288	&75.9	&92.3	&83.4	&72.2	&82.1 &29.8 &22.1\\
&2021	&\cite{liu2021polarized}	&no	 &HRNet-W48	        &384$\times$288	&\color{red}79.5	&93.6	& \color{blue}85.9	& \color{red}76.3	&\color{red}84.3 &35.4 &70.1\\
\multirow{-15}{*}{T} &2022	&\cite{ma2022ppt}	&no	 &Transformer	        &256$\times$192	&75.2	&89.8	&81.7	&  71.7	&82.1  &20.8 &8.7\\ \hline
&2017	&\cite{newell2017associative}	&no	&Hourglass 	                &512$\times$512	        &65.5	&86.8	&72.3	&60.6	&72.6 &- &-\\
&2018	&\cite{papandreou2018personlab}	&no	&ResNet-152	  	  &1401$\times$1401	        &68.7	&89.0	&75.4	&64.1	&75.5 &68.7 &405.5\\
&2019	&\cite{tian2019directpose}	&no	&ResNet-101  	                &800$\times$800	        &64.8	&87.8	&71.1	&60.4	&71.5 &- &-\\
&2020	&\cite{jin2020differentiable}	&no	&Hourglass	                &512$\times$512        &67.6	&85.1	&73.7	&62.7	&74.6 &- &-\\
&2020	&\cite{cheng2020higherhrnet}	&no	&HRNet-W48 	                &640$\times$640        &70.5	&89.3	&77.2	&66.6	&75.8 &63.8 &154.3\\
&2021	&\cite{luo2021rethinking}	&no	&HRNet-W48	                &640$\times$640        &72.0	&90.7	&78.8	&67.8	&77.7 &63.8 &154.6\\
\multirow{-9}{*}{B}&2022	&\cite{wang2022regularizing}	&no	&HRNet-W32	                &640$\times$640        &72.8	&91.2	&79.9	&68.3	&79.3 &- &-\\ \hline

\end{tabular}
}
\label{COCO, using AP}
\vspace{-10pt}
\end{table}

\begin{table}[!htbp]
\centering
\setlength\tabcolsep{25.0pt} 
  \caption{\footnotesize{Comparison of different 2D video-based HPE methods on the PoseTrack2017 test set and PoseTrack2018 test set}}
  \vspace{-5pt}
  \resizebox{\linewidth}{!}{
\begin{tabular}{|ccccccccccc|}
\hline
\multicolumn{11}{|c|}{PoseTrack2017} \\ \hline
\multicolumn{1}{|c|}{Year} & \multicolumn{1}{c|}{Method} & \multicolumn{1}{c|}{Backbone} & \multicolumn{1}{c|}{Head} & \multicolumn{1}{c|}{Shoulder} & \multicolumn{1}{c|}{Elbow} & \multicolumn{1}{c|}{Wrist} & \multicolumn{1}{c|}{Hip}  & \multicolumn{1}{c|}{Knee} & \multicolumn{1}{c|}{Ankle} & Total \\ \hline
\multicolumn{1}{|c|}{2018} & \multicolumn{1}{c|}{\cite{doering2018joint}}   & \multicolumn{1}{c|}{VGG}    & \multicolumn{1}{c|}{-}    & \multicolumn{1}{c|}{-}        & \multicolumn{1}{c|}{-}     & \multicolumn{1}{c|}{53.1}  & \multicolumn{1}{c|}{-}    & \multicolumn{1}{c|}{-}    & \multicolumn{1}{c|}{50.4}  & 63.4  \\ \hline
\multicolumn{1}{|c|}{2019} & \multicolumn{1}{c|}{\cite{bertasius2019learning}}  & \multicolumn{1}{c|}{HRNet-W48} & \multicolumn{1}{c|}{79.5} & \multicolumn{1}{c|}{84.3}     & \multicolumn{1}{c|}{80.1}  & \multicolumn{1}{c|}{75.8}  & \multicolumn{1}{c|}{77.6} & \multicolumn{1}{c|}{76.8} & \multicolumn{1}{c|}{70.8}  & 77.9  \\ \hline
\multicolumn{1}{|c|}{2020} & \multicolumn{1}{c|}{\cite{snower202015}}     & \multicolumn{1}{c|}{Transformer}    & \multicolumn{1}{c|}{-}    & \multicolumn{1}{c|}{-}        & \multicolumn{1}{c|}{-}     & \multicolumn{1}{c|}{71.9}  & \multicolumn{1}{c|}{-}    & \multicolumn{1}{c|}{-}    & \multicolumn{1}{c|}{65.0}    & 74.0    \\ \hline
\multicolumn{1}{|c|}{2021} & \multicolumn{1}{c|}{\cite{liu2021deep}}          & \multicolumn{1}{c|}{HRNet-W48}  & \multicolumn{1}{c|}{\color{blue}84.3} & \multicolumn{1}{c|}{\color{blue}84.9}     & \multicolumn{1}{c|}{\color{blue}80.5}  & \multicolumn{1}{c|}{\color{blue}76.1}  & \multicolumn{1}{c|}{\color{blue}77.9} & \multicolumn{1}{c|}{\color{blue}77.1} & \multicolumn{1}{c|}{\color{blue}71.2}  & \color{blue}79.2  \\ \hline
\multicolumn{1}{|c|}{2022} & \multicolumn{1}{c|}{\cite{liu2022temporal}}     & \multicolumn{1}{c|}{HRNet-W48} & \multicolumn{1}{c|}{\color{red}86.1} & \multicolumn{1}{c|}{\color{red}86.1}     & \multicolumn{1}{c|}{\color{red}81.8}  & \multicolumn{1}{c|}{\color{red}77.4}  & \multicolumn{1}{c|}{\color{red}79.5} & \multicolumn{1}{c|}{\color{red}79.1} & \multicolumn{1}{c|}{\color{red}73.6}  & \color{red}80.9  \\ \hline
\multicolumn{11}{|c|}{PoseTrack2018}\\ \hline
\multicolumn{1}{|c|}{2018} & \multicolumn{1}{c|}{\cite{guo2018multi}}    & \multicolumn{1}{c|}{ResNet-152}       & \multicolumn{1}{c|}{-}    & \multicolumn{1}{c|}{-}        & \multicolumn{1}{c|}{-}     & \multicolumn{1}{c|}{74.5}  & \multicolumn{1}{c|}{-}    & \multicolumn{1}{c|}{-}    & \multicolumn{1}{c|}{69.0}    & 76.4  \\ \hline
\multicolumn{1}{|c|}{2019} & \multicolumn{1}{c|}{\cite{bertasius2019learning}} & \multicolumn{1}{c|}{HRNet-W48} & \multicolumn{1}{c|}{78.9} & \multicolumn{1}{c|}{\color{blue}84.4}     & \multicolumn{1}{c|}{\color{blue}80.9}  & \multicolumn{1}{c|}{76.8}  & \multicolumn{1}{c|}{75.6} & \multicolumn{1}{c|}{77.5} & \multicolumn{1}{c|}{71.8}  & 78.0    \\ \hline
\multicolumn{1}{|c|}{2021} & \multicolumn{1}{c|}{\cite{liu2021deep}}   & \multicolumn{1}{c|}{HRNet-W48}         & \multicolumn{1}{c|}{\color{blue}82.8} & \multicolumn{1}{c|}{84.0}       & \multicolumn{1}{c|}{80.8}  & \multicolumn{1}{c|}{\color{blue}77.2}  & \multicolumn{1}{c|}{\color{blue}76.1} & \multicolumn{1}{c|}{\color{blue}77.6} & \multicolumn{1}{c|}{\color{blue}72.3}  & \color{blue}79.0    \\ \hline
\multicolumn{1}{|c|}{2022} & \multicolumn{1}{c|}{\cite{liu2022temporal}}   & \multicolumn{1}{c|}{HRNet-W48}     & \multicolumn{1}{c|}{\color{red}83.6} & \multicolumn{1}{c|}{\color{red}84.5}     & \multicolumn{1}{c|}{\color{red}81.4}  & \multicolumn{1}{c|}{\color{red}77.9}  & \multicolumn{1}{c|}{\color{red}76.8} & \multicolumn{1}{c|}{\color{red}78.3} & \multicolumn{1}{c|}{\color{red}72.9}  & \color{red}79.6  \\ \hline
\end{tabular}
}
\label{posetrack}
\vspace{-10pt}
\end{table}


\begin{table*}[!htbp]
\tiny
\centering
  \caption{\footnotesize{Datasets for 3D HPE.}}
  \vspace{-5pt}
  \scalebox{0.78}
{
\begin{tabular}{|c|c|c|c|c|c|c|c|c|}
\hline
{Dataset} & {Year} & {Capture system} & {Environment} & {Size}               & Single person & Multi-person & Single view & Multi-view              
\\ \hline
HumanEva \cite{HumanEva}                       & 2010                  & Marker-based MoCap                 & Indoor                  & 6 subject, 7 actions, 40k frames         & Yes                  & No                   & Yes                  & Yes                  \\ \hline
Human3.6M  \cite{Human3.6M}              & 2014                  & Marker-based MoCap              & Indoor                       & 11 subjects, 17 actions, 3.6M frames & Yes                                                                       & No                                                                       & Yes                                                                     & Yes                                                                    \\ \hline
CMU  Panoptic \cite{CMU-Panoptic}           & 2016                  & Marker-less MoCap               & Indoor                       & 8 subjects, 1.5M frames             & Yes                                                                       & Yes                                                                      & Yes                                                                     & Yes                                                                     \\ \hline
MPI-INF-3DHP \cite{MPI-INF-3DHP}             & 2017                  & Marker-less MoCap               & Indoor and outdoor           & 8 subjects, 8 actions, 1.3M frames  & Yes                                                                       & No                                                                       & Yes                                                                     & Yes                                                                    \\ \hline
TotalCapture \cite{TotalCapture}                       & 2017                  & Marker-based MoCap with IMUs
                 & Indoor                  & 5 subjects, 5 actions, 1.9M frames
         & Yes                  & No                   & Yes                  & Yes                  \\ \hline
3DPW \cite{Marcard_2018_ECCV}                    & 2018                  & Hand-held cameras with IMUs     & Indoor and outdoor           & 7 subjects, 51k frames              & Yes                                                                       & Yes                                                                      & Yes                                                                     & No                                                                     \\ \hline
MuPoTS-3D \cite{Mehta20183DV}               & 2018                  & Marker-less MoCap               & Indoor and outdoor           & 8 subjects, 8k frames               & Yes                                                                       & Yes                                                                      & Yes                                                                     & Yes                                                                    \\ \hline
AMASS \cite{AMASS2019}                   & 2019                  & Marker-based MoCap              & Indoor and outdoor           & 300 subjects, 9M frames             & Yes                                                                       & No                                                                       & Yes                                                                     & Yes                                                                    \\ \hline
NBA2K \cite{nba2k_ECCV2020}                   & 2020                  & NBA2K19 game engine             & Indoor                       & 27 subjects, 27k poses               & Yes                                                                       & No                                                                       & Yes                                                                     & No                                                                     \\ \hline
GTA-IM \cite{GTA-IM-ECCV2020}                  & 2020                  & GTA game engine                 & Indoor                       & 1M frames                           & Yes                                                                       & No                                                                       & Yes                                                                     & No                                                                     \\
\hline
Occlusion-Person \cite{zhang2020adafuse}                  & 2020                  &Unreal Engine 4 game engine                 & Indoor                       & 73k frames                           & Yes                                                                       & No                                                                       & Yes                                                                     & Yes                                                                     \\\hline
\end{tabular}}
\label{tab:3Ddatasets}
\vspace{-12pt}
\end{table*}

\subsection{Performance Comparison of 2D HPE Methods}
\textbf{Single-person 2D HPE:} Table \ref{MPII dataset on testing, using PCKh@0.5}  shows the comparison results for different 2D single-person HPE methods on the MPII dataset using PCKh@0.5 measure. Although both heatmap-based and regression-based methods have impressive results, they have their own limitations in 2D single-person HPE. Regression methods can learn a nonlinear mapping from input images to keypoint coordinates with an end-to-end framework, which offers a fast learning  paradigm and a sub-pixel level prediction accuracy. However, they usually give sub-optimal solutions \cite{luvizon2019human} due to the highly nonlinear problem. 
Heatmap-based methods outperform regression-based approaches and are more widely used in 2D HPE\cite{li2019rethinking}\cite{cai2020learning}\cite{luvizon2019human} since (1) the probabilistic prediction of each pixel in a heatmap can improve the accuracy of locating the keypoints, and (2) heatmaps provide richer supervision information by preserving the spatial location information. However, the precision of the predicted keypoints is dependent on the resolution of heatmaps. The computational cost and memory footprint are significantly increased when using high-resolution heatmaps\cite{sun2019deep}.   

\textbf{Multi-person 2D HPE:} 
Table \ref{COCO, using AP} presents the experimental results of different 2D HPE methods on the test-dev set of the COCO dataset, together with a summary of  the  experiment  settings  (extra data, backbones in models, input image size) and AP scores for each approach. The comparison experiments highlight the significant results of both top-down and bottom-up methods in multi-person HPE. Presumably, the top-down pipeline yields better results because it first detects each individual from the image using detection methods, then predicts the locations of keypoints using single-person HPE approaches. In this case, the keypoint estimation for each detected person is made easier, as the background is largely removed. But on the other hand, bottom-up methods are generally faster than top-down methods, because they directly detect all the keypoints and group them into individual poses using keypoint association strategies such as affinity linking \cite{cao2017realtime}, associative embedding \cite{newell2017associative}, and pixel-wise keypoint regression \cite{zhou2019objects}. Besides the image-based methods listed above, Table \ref{posetrack} also illustrates the comparisons of the recent video-based works on PoseTrack2017 and PoseTrack2018 datasets. The detailed results of the test sets are summarized.

\subsection{Datasets for 3D HPE}

In contrast to numerous 2D human pose datasets with high-quality annotation, acquiring accurate 3D annotation for 3D HPE datasets is a challenging task that requires motion capture systems such as MoCap and wearable IMUs. 
Due to the page limit, we only review several widely used large-scale 3D pose datasets for deep learning-based 3D HPE in the following.


\href{http://vision.imar.ro/human3.6m/}{\textbf{Human3.6M}} \cite{Human3.6M} is the most widely used indoor dataset for 3D HPE from monocular images and videos. There are 11 professional actors performing 17 activities 
from 4 different views in an indoor laboratory environment. This dataset contains 3.6 million 3D human poses with 3D ground truth annotation captured by accurate marker-based MoCap systems.  Protocol \#1 uses images of subjects S1, S5, S6, and S7 for training, and images of subjects S9 and S11 for testing.
\href{http://gvv.mpi-inf.mpg.de/projects/SingleShotMultiPerson/}{\textbf{MuPoTS-3D}} \cite{Mehta20183DV} is a multi-person 3D test set and its ground-truth 3D poses were captured by a multi-view marker-less MoCap system containing 20  real-world scenes (5 indoor and 15 outdoor). There are challenging samples with occlusions, drastic illumination changes, and lens flares in some of the outdoor footage. More than 8,000 frames were collected in the 20 sequences by 8 subjects.

Readers are referred to the original references for details about other datasets including  \textbf{MPI-INF-3DHP}\cite{MPI-INF-3DHP}, \textbf{HumanEva} \cite{HumanEva}, \textbf{CMU Panoptic Dataset} \cite{CMU-Panoptic}, \textbf{TotalCapture} \cite{TotalCapture},\textbf{MuCo-3DHP Dataset} \cite{Mehta20183DV}, \textbf{3DPW} \cite{Marcard_2018_ECCV}, \textbf{AMASS} \cite{AMASS2019}, \textbf{NBA2K} \cite{nba2k_ECCV2020}, \textbf{GTA-IM} \cite{GTA-IM-ECCV2020}, and \textbf{Occlusion-Person} \cite{zhang2020adafuse}. A summary of these datasets is shown in Table \ref{tab:3Ddatasets}.

\begin{table}[]
\tiny
\centering
\setlength\tabcolsep{5.0pt} 
\caption{\footnotesize{Comparison of different 3D single-view single-person HPE approaches on the Human3.6M dataset (Protocol 1). In skeleton-only approaches, ``Direct'' indicates the methods directly estimating 3D pose without 2D pose representation. ``Lifting'' denotes the methods lifting the 2D pose representation to the 3D space. The reported total parameters (Params) and FLOPs for 2D-3D lifting methods are computed without including the params and FLOPs of the external 2D pose detector. }}
\vspace{-10pt}
\resizebox{\linewidth}{!}
  {
\begin{tabular}{|ccccccc|cccccccccccccccc|c|}
\hline
\multicolumn{7}{|c|}{\textbf{Skeleton-only methods}}                                                                      & \multicolumn{16}{c|}{MPJPE}                                                                                                                                                                                                                                                                                                                                                                                                                                                                   & PA-MPJPE                    \\ \hline
\multicolumn{1}{|c|}{Approaches}                 & Year & Methods & \multicolumn{1}{l}{2D pose detector} & Input & Params(M) & FLOPs(G)  & Dir.                        & Disc.                       & Eat.                        & Greet                       & Phone                       & Photo                       & Pose                        & Purch.                      & Sit                         & SitD.                       & Somke                       & Wait                        & WalkD.                      & Walk                        & WalkT.                      & Average                     & Average                     \\ \hline
\multicolumn{1}{|c|}{}                           & 2017 & \cite{pavlakos2017volumetric}     & -                                    & image  & - & -  & 67.4                        & 71.9                        & 66.7                        & 69.1                        & 72.0                        & 77.0                        & 65.0                        & 68.3                        & 83.7                        & 96.5                        & 71.7                        & 65.8                        & 74.9                        & 59.1                        & 63.2                        & 71.9                        & 51.9                        \\
\multicolumn{1}{|c|}{\multirow{-2}{*}{Direct}}   & 2018 & \cite{pavlakos2018ordinal}     & -                                    & image & - & - & 48.5                        & 54.4                        & 54.4                        & 52.0                        & 59.4                        & 65.3                        & 49.9                        & 52.9                        & 65.8                        & 71.1                        & 56.6                        & 52.9                        & 60.9                        & 44.7                        & 47.8                        & 56.2                        & 41.8                        \\ \hline
\multicolumn{1}{|c|}{}                           & 2019 & \cite{Li_2019_CVPR}
     & Hourglass                            & image  & - & - & 43.8                        & 48.6                        & 49.1                        & 49.8                        & 57.6                        & 61.5                        & 45.9                        & 48.3                        & 62.0                        & 73.4                        & 54.8                        & 50.6                        & 56.0                        & 43.4                        & 45.5                        & 52.7                        & 42.6                        \\
\multicolumn{1}{|c|}{}                           & 2019 & \cite{Zhao_2019_Semantic_Graph}
     & own design                           & image  & - & - & 47.3                        & 60.7                        & 51.4                        & 60.5                        & 61.1                        & {\color{blue} 49.9} & 47.3                        & 68.1                        & 86.2                        & {\color{red} 55.0} & 67.8                        & 61.0                        & {42.1} & 60.6                        & 45.3                        & 57.6                        & -                           \\
\multicolumn{1}{|c|}{}                           & 2021 & \cite{Zou_2021_mgcn}     & CPN                                  & image & 0.3 & -  & 45.4                        & 49.2                        & 45.7                        & 49.4                        & 50.4                        & 58.2                        & 47.9                        & 46.0                        & 57.5                        & 63.0                        & 49.7                        & 46.6                        & 52.2                        & 38.9                        & 40.8                        & 49.4                        & 39.1                        \\
\multicolumn{1}{|c|}{}                           & 2019 & \cite{pavllo2019}     & CPN                                  & Video & 17 & 0.03  & 45.2                        & 46.7                        & 43.3                        & 45.6                        & 48.1                        & 55.1                        & 44.6                        & 44.3                        & 57.3                        & 65.8                        & 47.1                        & 44.0                        & 49.0                        & 32.8                        & 33.9                        & 46.8                        & 36.5                        \\
\multicolumn{1}{|c|}{}                           & 2019 & \cite{Cai2019Spatial-Temporal}     & CPN                                  & Video & - & -  & 44.6                        & 47.4                        & 45.6                        & 48.8                        & 50.8                        & 59.0                        & 47.2                        & 43.9                        & 57.9                        & 61.9                        & 49.7                        & 46.6                        & 51.3                        & 37.1                        & 39.4                        & 48.8                        & 39.0                        \\
\multicolumn{1}{|c|}{}                           & 2020 & \cite{Liu_2020_CVPR}
     & CPN                                  & Video & 11 & -  & 41.8                        & 44.8                        & 41.1                        & 44.9                        & 47.4                        & 54.1                        & 43.4                        & 42.2                        & 56.2                        & 63.6                        & 45.3                        & 43.5                        & 45.3                        & {\color{blue} 31.3} & 32.2                        & 45.1                        & 35.6                        \\
\multicolumn{1}{|c|}{}                           & 2020 & \cite{zeng2020srnet_ECCV}
     & CPN                                  & Video & - & -  & 46.6                        & 47.1                        & 43.9                        & 41.6                        & 45.8                        & {\color{red}49.6} & 46.5                        & {\color{blue}40.0} & 53.4                        & 61.1                        & 46.1                        & {42.6} & 43.1                        & 31.5                        & 32.6                        & 44.8                        & -                           \\
\multicolumn{1}{|c|}{}                           & 2020 & \cite{wang2020motion}
     & HRNet                                & Video & - & -  & {\color{blue} 38.2} & {\color{blue} 41.0} & 45.9                        & {\color{red}39.7} & {\color{red}41.4} & 51.4                        & {41.6} & 41.4                        & {\color{blue} 52.0} & {\color{blue} 57.4} & {\color{red} 41.8} & 44.4                        & {\color{blue} 41.6} & 33.1                        & {\color{blue} 30.0} & {\color{blue} 42.6} & {\color{blue} 32.7} \\
\multicolumn{1}{|c|}{}                           & 2020 & \cite{chen2020anatomy}
     & CPN                                  & Video & 59 & 0.1  & 41.4                        & 43.5                        & { 40.1} & 42.9                        & 46.6                        & 51.9                        & 41.7                        & 42.3                        & 53.9                        & 60.2                        & 45.4                        & 41.7                        & 46.0                        & 31.5                        & 32.7                        & 44.1                        & 35.0                        \\
\multicolumn{1}{|c|}{}                           & 2021 & \cite{zheng20213d}
     & CPN                                  & Video & 10 & 1.4  & 41.5                        & 44.8                        & {\color{blue} 39.8} & 42.5                        & 46.5                        & 51.6                        & 42.1                        & 42.0                        & { 53.3} & 60.7                        & 45.5                        & 43.3                        & 46.1                        & 31.8                        & 32.2                        & 44.3                        & {34.6} \\
\multicolumn{1}{|c|}{} & 2022 & \cite{li2021mhformer}
     & CPN                                  & Video & 32 & 7.0  & {39.2} & {43.1} & 40.1                        & {\color{blue}40.9} & {44.9} & 51.2                        & {\color{blue}40.6} & {41.3} & 53.5                        & 60.3                        & {43.7} & {\color{blue}41.1} & 43.8                        & {\color{red}29.8} & {30.6} & {43.0} & -                           \\
\multicolumn{1}{|c|}{\multirow{-15}{*}{Lifting}} & 2022 & \cite{MixSTE}
     & CPN                                  & Video & 41 & 0.6   & {\color{red}37.6} & {\color{red}40.9} & \color{red}37.3                        & {\color{red}39.7} & {\color{blue}42.3} & \color{blue}49.9                        & {\color{red} 40.1} & {\color{red}39.8} & \color{red}51.7                        & \color{red}55.0                        & {\color{blue}42.1} & {\color{red}39.8} & \color{red}41.0                        & {\color{red}27.9} & {\color{red}27.9} & {\color{red}40.9} & \color{red}32.6                           \\ \hline
\multicolumn{7}{|c|}{\textbf{Human mesh recovery methods}}                                                                & \multicolumn{16}{c|}{MPJPE}                                                                                                                                                                                                                                                                                                                                                                                                                                                                   & PA-MPJPE                    \\ \hline
\multicolumn{1}{|c|}{output}                     & Year & Methods & Model                                & Input & Params(M) & FLOPs(G)  & Dir.                        & Disc.                       & Eat.                        & Greet                       & Phone                       & Photo                       & Pose                        & Purch.                      & Sit                         & SitD.                       & Somke                       & Wait                        & WalkD.                      & Walk                        & WalkT.                      & Average                     & Average                     \\ \hline
\multicolumn{1}{|c|}{}                           & 2019 & \cite{kolotouros2019cmr}
     & SMPL                                 & Image & - & -  & -                           & -                           & -                           & -                           & -                           & -                           & -                           & -                           & -                           & -                           & -                           & -                           & -                           & -                           & -                           & -                           & 50.1                        \\
\multicolumn{1}{|c|}{}                           & 2019 & \cite{Kolotouros2019SPIN}
     & SMPL                                 & Image & - & -  & -                           & -                           & -                           & -                           & -                           & -                           & -                           & -                           & -                           & -                           & -                           & -                           & -                           & -                           & -                           & -                           & 41.1                        \\
\multicolumn{1}{|c|}{}                           & 2019 & \cite{xiang2019monocular}
     & Adam                                 & Image & - & -   & -                           & -                           & -                           & -                           & -                           & -                           & -                           & -                           & -                           & -                           & -                           & -                           & -                           & -                           & -                           & 58.3                        & -                           \\
\multicolumn{1}{|c|}{}                           & 2020 & \cite{Choi_2020_ECCV_Pose2Mesh}
     & -                                 & Image & 140 & 11   & -                           & -                           & -                           & -                           & -                           & -                           & -                           & -                           & -                           & -                           & -                           & -                           & -                           & -                           & -                           & 64.9                        & 47.0                        \\
\multicolumn{1}{|c|}{}                           & 2021 & \cite{pymaf2021}
     & SMPL                                 & Image & 45.2 & 11  & -                           & -                           & -                           & -                           & -                           & -                           & -                           & -                           & -                           & -                           & -                           & -                           & -                           & -                           & -                           & 57.7                        & 40.5                        \\
\multicolumn{1}{|c|}{}                           & 2021 & \cite{lin2021metro}
     & SMPL                                 & Image & 230 & 57  & -                           & -                           & -                           & -                           & -                           & -                           & -                           & -                           & -                           & -                           & -                           & -                           & -                           & -                           & -                           & {54.0} & {36.7} \\
\multicolumn{1}{|c|}{}                           & 2021 & \cite{lin2021_mesh_graphormer}
     & SMPL                                 & Image & 230 & 57  & -                           & -                           & -                           & -                           & -                           & -                           & -                           & -                           & -                           & -                           & -                           & -                           & -                           & -                           & -                           & {\color{red} 51.2} & {\color{blue} 34.5} \\
\multicolumn{1}{|c|}{}                           & 2022 & \cite{cho2022FastMETRO}
     & SMPL                                 & Image & 49 & 16  & -                           & -                           & -                           & -                           & -                           & -                           & -                           & -                           & -                           & -                           & -                           & -                           & -                           & -                           & -                           & {\color{blue}52.2} & {\color{red}33.7} \\     \cline{2-24} 
\multicolumn{1}{|c|}{}                           & 2019 & \cite{Arnab_2019_CVPR}
     & SMPL                                 & Video & - & -  & -                           & -                           & -                           & -                           & -                           & -                           & -                           & -                           & -                           & -                           & -                           & -                           & -                           & -                           & -                           & 77.8                        & 54.3                        \\
\multicolumn{1}{|c|}{}                           & 2020 & \cite{kocabas2020vibe}
     & SMPL                                 & Video & 59 & 10  & -                           & -                           & -                           & -                           & -                           & -                           & -                           & -                           & -                           & -                           & -                           & -                           & -                           & -                           & -                           & 65.6                        & 41.4                        \\
\multicolumn{1}{|c|}{\multirow{-11}{*}{Mesh}}    & 2021 & \cite{TCMR_Choi_2021}
     & SMPL                                 & Video & 123 & 10  & -                           & -                           & -                           & -                           & -                           & -                           & -                           & -                           & -                           & -                           & -                           & -                           & -                           & -                           & -                           & 62.3                        & 41.1                        \\ \hline
\end{tabular}
}
\label{tab: single-person Human 3.6 datasets performance}
\vspace{-15pt}
\end{table}

\subsection{Evaluation Metrics for 3D HPE}

\textbf{MPJPE} (Mean Per Joint Position Error) is the most widely used metric to evaluate the performance of 3D HPE. MPJPE is computed by using the Euclidean distance between the estimated 3D joints and the ground truth positions 
as follows: 
\begin{align}
\small
    {MPJPE} =\frac{1}{N} \sum_{i=1}^N \|J_i - J_i^* \|_2,
\end{align}
where $N$ is the number of joints, $J_i$ and $J^*_i$ are the ground truth position and the estimated position of the $i_{th}$ joint.

\textbf{PA-MPJPE}, also called Reconstruction Error, is the MPJPE after rigid alignment by post-processing between the estimated pose and the ground truth pose. 

\textbf{NMPJPE} is defined as the MPJPE after normalizing the predicted positions in scale to the reference \cite{Rhodin_2018_CVPR}.

\textbf{MPVE} (Mean Per Vertex Error)  \cite{pavlakos2018humanshape} measures
the Euclidean distances between the ground truth vertices and the predicted vertices: 
\begin{align}
\small
    {MPVE} =\frac{1}{N} \sum_{i=1}^N \|V_i - V_i^* \|_2,
\end{align}
where $N$ is the number of vertices, $V$ is the ground truth vertices, and $V^*$ is the estimated vertices. 

\textbf{3DPCK} is a 3D extended version of the Percentage of Correct Keypoints (PCK) metric used in 2D HPE evaluation. An estimated joint is considered as correct if the distance between the estimation and the ground truth is within a certain threshold. Generally, the threshold is set to 150$mm$. 

\begin{table}[H]
\begin{minipage}{0.45\linewidth}
\tiny
\vspace{-7pt}
\centering
\setlength\tabcolsep{3.0pt} 
\renewcommand\arraystretch{0.9}
\caption{\footnotesize{Comparison of different 3D single-view multi-person HPE approaches on the MuPoTS-3D dataset. The reported fps is taken from \cite{wang2022distribution}.}}
\vspace{-7pt}
\begin{tabular}{|c|c|c|c|c|c|}
\hline
\multicolumn{6}{|c|}{MuPoTS-3D}                                                                                                                                   \\ \hline
                            &                        & \multicolumn{1}{c|}{}                          & \multicolumn{2}{c|}{3DPCK $\uparrow$}  &\multirow{2}{*}{fps}                              \\ \cline{4-5} 
\multirow{-2}{*}{}  & \multirow{-2}{*}{Year} & \multicolumn{1}{c|}{\multirow{-2}{*}{Method}} & All people                         & Matched people   &                     \\ \hline
                            & 2019                   & \cite{Moon_2019_ICCV_3DMPPE}                                            & { 81.8}    & {\color{red} 82.5} & 9.3   \\
                            & 2020                   & \cite{Jiang_2020_CVPR}                                            & { 69.1}    & {72.2}   & - \\
                            & 2020                   & \cite{Li_ECCV2020_HMOR}                                            & { 82.0}    & {-}  & -  \\
\multirow{-5}{*}{Top down}  & 2021                   & \cite{Cheng_Wang_Yang_Tan_2021}                                            & {\color{blue}  87.5}                        & -                   & -     \\ \hline
                            & 2018                   & \cite{Mehta20183DV}                                            & 65.0                        & 69.8       & -                 \\
                            & 2019                   & \cite{mehtaxnect}                                            & 70.4                        &   -          & -                \\
                            & 2020                   & \cite{Benzine_2020_CVPR}                                            & 72.0                        & -                       & -     \\
\multirow{-4}{*}{Bottom up} & 2020                   & \cite{Zhen_ECCV2020_SMAP}                                            & {73.5} & {\color{blue} 80.5} & 9.3\\
\hline
 & 2021                   & \cite{Cheng_2021_TDBU}                                            & {\color{red} 89.6} & {-} & -\\
\multirow{-2}{*}{Integrated} & 2022                   & \cite{wang2022distribution}                                            & {82.7} & {-} & 13.3\\
\hline
\end{tabular}
\vspace{-5pt}
\label{tab: multi-person MuPoTS-3D datasets performance}
\end{minipage}
\quad
\begin{minipage}{0.45\linewidth}  
\vspace{-7pt}
\centering
\tiny
\setlength\tabcolsep{1.0pt} 
\renewcommand\arraystretch{0.9}
\caption{\footnotesize{Comparison of different 3D multi-view HPE approaches on the Human3.6M dataset. }}
\vspace{-7pt}
\begin{tabular}{|c|c|c|c|c|c|c|}
\hline
\multicolumn{7}{|c|}{Human3.6M}                                                                                                                                                                                                                                                 \\ \hline
                      &                           &                                                                                & \multicolumn{2}{c|}{Protocol 1}                                                                             & Protocol 1            &  \multirow{3}{*}{FLOPs(G)}      \\ \cline{4-6} 
                      &                           &                                                                                &                             &                                                                               &                         &    \\
\multirow{-3}{*}{Year} & \multirow{-3}{*}{Method} & \multirow{-3}{*}{\begin{tabular}[c]{@{}c@{}}Use extra\\  3D data\end{tabular}} & \multirow{-2}{*}{MPJPE $\downarrow$}     & \multirow{-2}{*}{\begin{tabular}[c]{@{}c@{}}Normalized\\  MPJPE $\downarrow$\end{tabular}} & \multirow{-2}{*}{PA-MPJPE $\downarrow$}  &   \\ \hline
2019                   & \cite{Liang_2019_ICCV}                       & Yes                                                                            & 79.9                        & -                                                                             & {\color{red} 45.1}  & -      \\
2019                   & \cite{Kocabas_2019_CVPR}                       & No                                                                             & 60.6                        & {\color{red} 60.0}                                                      & {\color{blue} 47.5} & -   \\
2019                   & \cite{Qiu_2019_ICCV}                       & No                                                                             & 31.2                        & -                                                                             & -                    & 55         \\
2019                   & \cite{Qiu_2019_ICCV}                       & Yes                                                                            &  26.2    & -                                                                             & -                    & 55          \\
2020                   & \cite{Remelli_2020_CVPR}                       & No                                                                             & 30.2                        & -                                                                             & -                      & -        \\
2020                   & \cite{Xie_2020_CVPR}                      & No                                                                             & {29.3} & -                                                                             & -                      & -       \\
2020                   & \cite{zhang2020adafuse}                      & Yes                                                                             & {\color{blue} 19.5} & -                                                                             & -                   & -          \\
2021                   & \cite{ma2021transfusion}                       & No                                                                             & 25.8                        & -                                                                             & -             & 50                \\
2021                   & \cite{reddy2021tessetrack}                      & Yes                                                                             & {\color{red} 18.7} & -                                                                             & -                & -             \\
2022                   & \cite{ma2022ppt}                      & Yes                                                                             & {24.4} & -                                                                             & -                       & 15       \\

\hline
\end{tabular}
\label{tab: multi-view Human3.6M dataset performance}
\vspace{-5pt}
\end{minipage}
\end{table}

\textbf{Summary.} As pointed out by Ji et al. \cite{JI-3DHPE-Survey}, low MPJPE does not always indicate an accurate pose estimation as it depends on the predicted scale of human shape and skeleton. Although 3DPCK is more robust to incorrect joints, it cannot evaluate the precision of correct joints. 
Existing metrics are designed to evaluate the precision of an estimated pose in a single frame. However, the temporal consistency and smoothness of the reconstructed human pose cannot be examined over continuous frames by existing evaluation metrics. Designing frame-level evaluation metrics that can evaluate 3D HPE performance with temporal consistency and smoothness remains an open problem.

\subsection{Performance Comparison of 3D HPE Methods}

\textbf{Single-view single-person}: In Table \ref{tab: single-person Human 3.6 datasets performance}, it is seen that most 3D single-view single-person HPE methods estimate 3D human pose with remarkable precision on the Human3.6M dataset. However, despite the fact that the Human3.6M dataset has a large size of training and testing data, it only contains 11 actors performing 17 activities in lab environments. When estimating 3D pose on in-the-wild data with more complex scenarios, the performance of these methods degrades quickly. Estimating 3D pose from videos can achieve better performance than from a single image because the temporal consistency is preserved.  

For skeleton-only methods, 2D-to-3D lifting approaches generally outperform direct estimation approaches due to the excellent performance of state-of-the-art 2D pose detectors.
Beyond estimated 3D coordinates of joints, a group of methods utilized volumetric models such as SMPL to recover human mesh. These methods still reported MPJPE of joints since the datasets do not provide the ground truth mesh vertices. However, the reported MPJPE is higher than those methods that only estimate 3D joints. One of the reasons is that these methods regressed both pose parameters and shape parameters, then fed in the volumetric model for mesh reconstruction, only evaluating MPJPE of joints can not show their strength. 

\textbf{Single-view multi-person}: Estimating 3D poses in the multi-person setting is a harder task than in single-person due to more severe occlusion. As shown in Tables \ref{tab: multi-person MuPoTS-3D datasets performance}, good progress has been made in the single-view multi-person HPE methods in recent years. The Top-Down methods perform better than Bottom-Up methods due to the state-of-the-art person detection methods and single-person HPE methods. On the other hand, Bottom-Up methods are more computationally and time efficient.

\textbf{Multi-view}: By comparing the results from Table \ref{tab: single-person Human 3.6 datasets performance} and Table \ref{tab: multi-view Human3.6M dataset performance}, it is evident that the performance (e.g., MPJPE under Protocol 1) of multi-view 3D HPE methods has improved compared to single-view 3D HPE methods using the same dataset and evaluation metric. Occlusion and depth ambiguity can be alleviated through the multi-view setting.


\section{Applications}\label{Application}
In this section, we review related works of exploring HPE for a few popular applications.

\textbf{Action recognition, prediction, detection, and tracking}: Pose information has been utilized as cues for various applications such as action recognition, prediction, detection, and tracking. Angelini et al. \cite{angelini2018actionxpose} proposed a real-time action recognition method using a pose-based algorithm. Yan et al. \cite{yan2018spatial} leveraged the dynamic skeleton modality of pose for action recognition. Markovitz et al. \cite{Markovitz_2020_CVPR} studied human pose graphs for anomaly detection of human actions in videos. Cao et al. \cite{GTA-IM-ECCV2020} used the predicted 3D pose for long-term human motion prediction. Sun et al. \cite{sun_viewinvariant_ECCV2020} proposed a view-invariant probabilistic pose embedding for video alignment. Hua et al. \cite{hua2023part} proposed an attention-based contrastive learning framework that integrates local similarity and global features for skeleton-based action
representations.

Pose-based video surveillance enjoys the advantage of preserving privacy by monitoring through pose and human mesh representation instead of human sensitive identities. Das et al. \cite{das2020vpn} embedded video with poses to identify activities of daily living for monitoring human behavior.

\textbf{Action correction and online coaching}:  
Some activities such as dancing, sporting, and professional training require precise human body control guidance. Normally personal trainers are responsible for pose correction and action guidance in a face-to-face manner. With the help of 3D HPE and action detection, AI personal trainers can make coaching more convenient by simply setting up cameras without a personal trainer present. Wang et al. \cite{Wang_AIcoach} designed an AI coaching system with a pose estimation module for personalized athletic training assistance.

\textbf{Clothes parsing}:
The e-commerce trends have brought about a noticeable impact on various aspects including clothes purchases. Clothing products in pictures can no longer satisfy customers' demands, and customers hope to see reliable appearances as they wear their selected clothes. Clothes parsing \cite{yu2019simulcap} \cite{Saito_2019_ICCV} and pose transfer \cite{li2019dense} make it possible by inferring the 3D appearance of a person wearing specific clothes. HPE can provide plausible human body regions for cloth parsing. Moreover, the recommendation system can be upgraded by evaluating appropriateness based on the inferred reliable 3D appearance of customers with selected items. Patel et al. \cite{patel20tailornet} achieved clothing prediction from 3D pose, shape, and garment style.

\textbf{Animation, movie, and gaming:}
Motion capture is the key component to present characters with complex movements and realistic physical interactions in industries of animation, movies, and gaming. Equipments are usually expensive and complicated to set up. HPE can provide realistic pose information while alleviating the demand for professional high-cost equipment \cite{Animation_posetransfer}\cite{animation_pose}.

\textbf{AR and VR}:
Augmented Reality (AR) technology aims to enhance the interactive experience of digital objects in the real-world environment. The objective of Virtual Reality (VR) technology is to provide an immersive experience for the users. AR and VR devices use human pose information as input to achieve their goals of different applications. A cartoon character can be generated in real-world scenes to replace a real person. Weng et al. \cite{Weng_2019_Photo_WakeUp} created 3D character animation from a single photo with the help of 3D pose estimation and human mesh recovery. Zhang et al. \cite{zhang2020vid2player} presented a pose-based system that converts broadcast tennis match videos into interactive and controllable video sprites. 

\textbf{Healthcare}:
HPE provides quantitative human motion information that physicians can use to diagnose some complex diseases, create rehabilitation training, and operate physical therapy. Lu et al. \cite{lu2020vision} designed a pose-based estimation system for assessing Parkinson’s disease motor severity. Gu et al. \cite{Physical_Therapy_gu} developed a pose-based physical therapy system that patients can be evaluated and advised at their homes. Furthermore, such a system can be established to detect abnormal actions and predict subsequent actions ahead of time. Alerts are sent immediately if the system determines that danger may occur. Chen et al. \cite{falldetection_chen} used HPE algorithms for fall detection monitoring in order to provide immediate assistance. Also, HPE methods can provide reliable posture labels of patients in hospital environments to augment research on neural correlates to natural behaviors \cite{chen2018patient}.

\section{Conclusion and Future Directions}\label{Discussion}
In this survey, we have presented a systematic overview of recent deep learning-based 2D and 3D HPE methods. A comprehensive taxonomy and performance comparison of these methods have been covered. 
Despite great success, there are still many
challenges as discussed in Sections \ref{2DHPE Summary} and \ref{3DHPE_Summary}. We further point out a few promising future directions to promote advances in HPE research. 
\setlist{nolistsep}
\begin{itemize}[noitemsep,leftmargin=*]


\item Domain adaptation for HPE. For some applications such as estimating human pose from infant images \cite{huang2020infant} or artwork collections\cite{VasePaintingsHPE}, there are not enough training data with ground truth annotations. Moreover, data for these applications exhibit different distributions from that of the standard pose datasets. HPE methods trained on existing standard datasets may not generalize well across different domains. The recent trend to alleviate the domain gap is utilizing GAN-based learning approaches. Nonetheless, how to effectively transfer the human pose knowledge to bridge domain gaps remains unaddressed.  

\item Human body models such as SMPL, SMPL-X, GHUM \& GHUML, and Adam are used to model human mesh representation. 
However, these models have a huge number of parameters. How to reduce the number of parameters while preserving the reconstructed mesh quality is an intriguing problem. Also, different people have various deformations of body shape. A more effective human body model may utilize other information such as BMI \cite{STAR2020} and silhouette \cite{li2020silhouette} for better generalization.

\item Most existing methods ignore human interaction with 3D scenes. There are strong human-scene relationship constraints that can be explored such as a human subject cannot be simultaneously present in the locations of other objects in the scene. The physical constraints with semantic cues can provide reliable and realistic 3D HPE.   

\item 3D HPE is employed in visual tracking and analysis. Existing 3D HPE from videos are not smooth and continuous. One reason is that the evaluation metrics such as MPJPE cannot evaluate the smoothness and the degree of realisticness. Appropriate frame-level evaluation metrics focusing on temporal consistency and motion smoothness should be developed. 

\item Existing well-trained networks pay less attention to resolution mismatch. The training data of HPE networks are usually high-resolution images or videos, which may lead to inaccurate estimation when predicting human pose from low-resolution input. The contrastive learning scheme \cite{simclr} (e.g., the original image and its low-resolution version as a positive pair) might be helpful for building resolution-aware HPE networks.

\item Deep neural networks in vision tasks are vulnerable to adversarial attacks. Imperceptible noise can significantly affect the performance of HPE. There are few works \cite{liu2019adversarial} \cite{Jain_2019_CVPRW_Robustness} that consider adversarial attacks for HPE. The study of defense against adversarial attacks can improve the robustness of HPE networks and facilitate real-world pose-based applications.

\item Human body parts may have different movement patterns and shapes due to the heterogeneity of the human body. A single shared network architecture may not be optimal for estimating all body parts with varying degrees of freedom. Neural Architecture Search (NAS) \cite{elsken2019neural} can search the optimal architecture for estimating each body part \cite{Chen_2020_NAS}. Also, NAS can be used for discovering efficient HPE network architectures to reduce the computational cost \cite{zhang2020efficientpose}. It is also worth exploring multi-objective NAS in HPE when multiple objectives (e.g, latency, accuracy, and energy consumption) have to be met.
\end{itemize}

\noindent \textbf{HPE workshops and challenges:} Finally, we list the HPE workshops and challenges (2017-2022) on our project page (\url{https://github.com/zczcwh/DL-HPE}) to facilitate research in this field.

\bibliographystyle{ACM-Reference-Format}
\bibliography{sample-base}

\end{document}